\documentclass[final]{cvpr}

\usepackage{times}
\usepackage{epsfig}
\usepackage{graphicx}
\usepackage{amsmath}
\usepackage{amssymb}

\usepackage{bm}
\usepackage{multirow}
\usepackage{booktabs}

\usepackage{epstopdf}
\newcommand{\bk}{{\bm{k}}}

\newcommand{\bq}{{\bm{q}}}

\newcommand{\bx}{{\bm{x}}}

\newcommand{\bz}{{\bm{z}}}

\newcommand{\bA}{{\bm{A}}}

\newcommand{\bM}{{\bm{M}}}

\newcommand{\bX}{{\bm{X}}}

\newcommand{\mO}{{\mathcal{O}}}

\newcommand{\mS}{{\mathcal{S}}}
\newcommand{\mT}{{\mathcal{T}}}

\newcommand{\mV}{{\mathcal{V}}}

\usepackage[pagebackref=true,breaklinks=true,colorlinks,bookmarks=false]{hyperref}

\def\cvprPaperID{1038} 
\def\confYear{CVPR 2021}
\begin{document}

\title{Learning Feature Aggregation for Deep 3D Morphable Models}
\author{Zhixiang Chen \\
Imperial College London \\
{\tt\small zhixiang.chen@imperial.ac.uk}
\and
Tae-Kyun Kim \\
Imperial College London and KAIST \\
{\tt\small tk.kim@imperial.ac.uk}
}

\maketitle

\begin{abstract}
3D morphable models are widely used for the shape representation of an object class in computer vision and graphics applications.
In this work, we focus on deep 3D morphable models that directly apply deep learning on  3D mesh data with a hierarchical structure to capture information at multiple scales.
While great efforts have been made to design the convolution operator, how to best aggregate vertex features across hierarchical levels deserves further attention.
In contrast to resorting to mesh decimation, we propose an attention based module to learn mapping matrices for better feature aggregation across hierarchical levels.
Specifically, the mapping matrices are generated by a compatibility function of the keys and queries.
The keys and queries are trainable variables, learned by optimizing the target objective, and shared by all data samples of the same object class.
Our proposed module can be used as a train-only drop-in replacement for the feature aggregation in existing architectures for both downsampling and upsampling.
Our experiments show that through the end-to-end training of the mapping matrices, we achieve state-of-the-art results on a variety of 3D shape datasets in comparison to existing morphable models.\footnote{\url{https://github.com/zxchen110/Deep3DMM/}}

\end{abstract}

\begin{figure*}[tb]
    \centering
    \includegraphics[width=.9\linewidth]{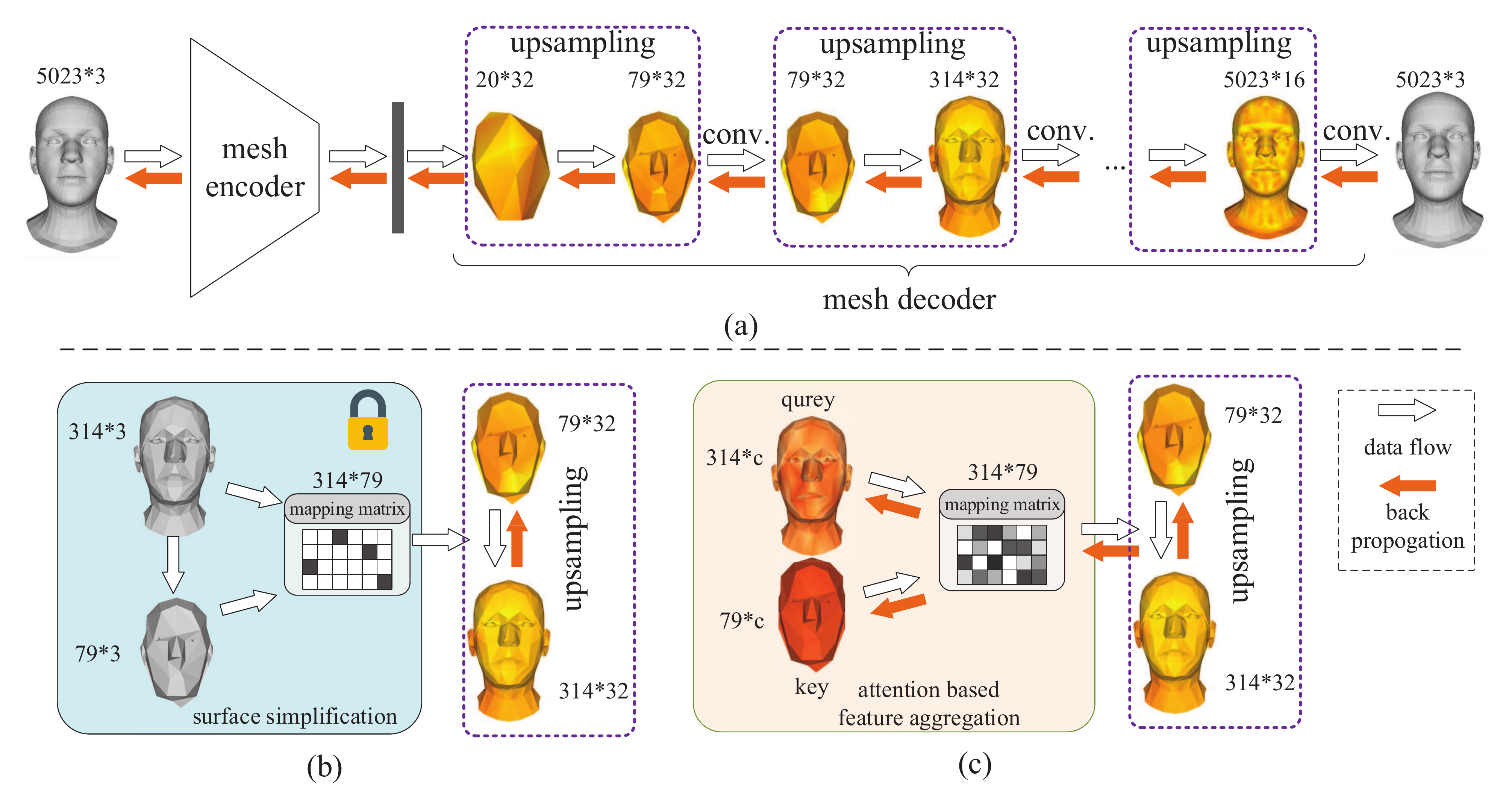}
    \caption{(a) Deep 3D morphable model with mesh encoder and decoder. The mesh encoder encodes a mesh as a compact latent representation, which the mesh decoder takes as input to recover the mesh with hierarchical upsampling and convolution operations. (b) A fixed mapping matrix generated by surface simplification is used for feature aggregation across hierarchical levels in~\cite{DBLP:conf/eccv/RanjanBSB18,DBLP:conf/iccv/BouritsasBPZB19}. (c) In this work, the mapping matrix for feature aggregation is generated by an attention based module and jointly learned with other components of the model}
    \label{fig:hp_frame_work}
\end{figure*}

\section{Introduction}
The 3D morphable models play a core role in many 3D applications, including identity recognition, shape retrieval, shape completion, animation and 3D reconstruction from 2D images~\cite{DBLP:conf/cvpr/TranHMPNM18,DBLP:conf/cvpr/TaigmanYRW14,DBLP:journals/tog/LiBBL017}.
3D morphable models encode raw 3D shapes into latent variables, from which the shapes can be reconstructed to some approximation by interacting with the 3D shape model.
In this work, we are particularly interested in 3D mesh shapes that share a common template and are already aligned to the template.
The commonly employed encoding approaches are the linear encoding of Principal Component Analysis~\cite{DBLP:conf/fgr/AmbergKV08,DBLP:journals/tog/LiBBL017,DBLP:conf/iccv/TewariZK0BPT17,DBLP:journals/tog/YangWSBM11} or manually defined blendshape~\cite{DBLP:journals/tog/ThiesZNVST15,DBLP:journals/tog/LiWP10,DBLP:journals/tog/BouazizWP13} for human face, skinned vertex based models like Skinned Multi-Person Linear Model~\cite{DBLP:journals/tog/LoperM0PB15} and hand model with articulated and non-rigid deformations~\cite{DBLP:journals/tog/0002TB17} for body and hand and non-linear encodings of autoencoder neural nets~\cite{DBLP:conf/cvpr/Tan0LX18}.
Compared to the classical linear models, the non-linear models, especially deep learning based models, offer the possibility to capture detailed deformations like wrinkles and surpass them in terms of generalization, compactness and specificity~\cite{DBLP:conf/ipmi/StynerRNZSTD03}.
\par
Recently, there is an emerging interest in generalizing CNNs to meshes with operations directly defined on meshes~\cite{DBLP:conf/eccv/RanjanBSB18,DBLP:conf/iccv/BouritsasBPZB19}.
Both isotropic~\cite{DBLP:conf/eccv/RanjanBSB18} and anisotropic~\cite{DBLP:conf/iccv/BouritsasBPZB19} convolutions have been introduced for meshes with a fixed topology.
Feature aggregation is defined for the downsampling and upsampling of vertex features and combined with convolutions to construct mesh autoencoders~\cite{DBLP:conf/eccv/RanjanBSB18}.
Despite the success of such models, the feature aggregation across hierarchical levels is determined by a preprocessing stage that maintains surface error approximations with quadric metrics~\cite{DBLP:conf/siggraph/GarlandH97} rather than by optimizing the learning objective.
Such procedure limits the representation power of the morphable model to capture fine grained deformations.
\par
In this work, we propose an attention based feature aggregation strategy to construct hierarchical representations for 3D meshes with a fixed topology (Fig.~\ref{fig:hp_frame_work}(a)).
Instead of using the precomputed matrices (Fig.~\ref{fig:hp_frame_work}(b)), 
we propose to learn the mapping matrices along with the convolutions in the network (Fig.~\ref{fig:hp_frame_work}(c)).
The attention based mapping module introduces keys and queries as trainable variables that are shared by all data samples in the dataset.
With keys and queries standing for the vertices at the preceding and succeeding levels respectively, the mapping matrices are derived by a compatibility function of the keys and queries.
This allows the receptive fields and aggregation weights to be simultaneously learned.
By varying the number of keys and queries, this module can be used for either downsampling or upsampling.
Since only mapping matrices are needed for inference, we can detach the attention based mapping module once training is finished to avoid additional cost at inference stage.
We evaluate our method on the reconstruction task which serves as a fundamental testbed for further applications.
We quantitatively and qualitatively show the results on three 3D human shape datasets: faces, bodies, and hands.
As a drop-in replacement for existing feature aggregation method, our method boosts the performance of existing models by a large margin, in combination with either isotropic or anisotropic convolution operators.

\section{Related work}

\subsection{3D Morphable Models}
3D Morphable Models represent the shape of objects as 3D meshes that can be deformed to match a particular instance of that object class. 
These meshes are in dense correspondence with a shared template of the object class.
Existing models take the statistical information of a set of representative examples of the class as prior knowledge to model shapes.
The most well studied models are on human face~\cite{DBLP:journals/tog/EggerSTWZBBBKRT20,DBLP:journals/tog/LiBBL017}, body~\cite{DBLP:journals/tog/LoperM0PB15,DBLP:conf/eccv/OsmanBB20} and hand~\cite{DBLP:journals/tog/0002TB17}.
Principal Component Analysis is the most commonly used approach for statistical modeling and has been used to model faces~\cite{DBLP:conf/siggraph/BlanzV99,DBLP:conf/avss/PaysanKARV09,DBLP:journals/tvcg/CaoWZTZ14,DBLP:journals/ijcv/BoothRPDZ18,DBLP:conf/cvpr/Ploumpis0PSZ19}.
SMPL~\cite{DBLP:journals/tog/LoperM0PB15,DBLP:conf/eccv/OsmanBB20} and MANO~\cite{DBLP:journals/tog/0002TB17} are the most well know models for body and hand.
These linear models filter out the high frequency signal of the shapes, thereby losing details, and require class specific joint localisation for shape deformation.
\par
Deep 3D morphable models apply 3D deep learning to model the nonlinear shape variations and are not limited to a specific class of shapes.
Litany \emph{et al.}~\cite{DBLP:conf/cvpr/LitanyBBM18} stack several graph convolutional layers and average the features of all vertices to obtain the mesh representation.
Ranjan \emph{et al.}~\cite{DBLP:conf/eccv/RanjanBSB18} propose a mesh autoencoder called CoMA with fast localised convolutional filters~\cite{DBLP:conf/nips/DefferrardBV16} and mesh downsampling and upseampling layer to support multi-scale hierarchical structure.
Bouritsas \emph{et al.}~\cite{DBLP:conf/iccv/BouritsasBPZB19} propose to replace the isotropic convolutions with anisotropic spiral convolutions.
Vertex-wise weighted convolutions are further proposed in~\cite{DBLP:conf/aaai/GaoZZYYY21,DBLP:conf/nips/ZhouWLCYSLS20} at the cost of increased number of parameters.
Tretschk \emph{et al.}~\cite{DBLP:conf/eccv/TretschkTZGT20} add an embedded deformation layer to the mesh autoencoder. 
Feature aggregation in most of these works adopts direct selection for downsampling and barycentric coordinates for upsampling. 
The aggregation functions are fixed and determined by quadric mesh simplification rather than minimizing reconstruction error.
Although the aggregation weights are learned from data in~\cite{DBLP:conf/nips/ZhouWLCYSLS20}, the aggregation is still performed over fixed local neigborhood.
In contrast, we propose to learn both the aggregation weights and receptive fields.
Furthermore, rather than directly setting the weights as trainable parameters, we generate the  weights by an attention based module to avoid over-parameterization.

\subsection{Geometric deep learning on meshes}
Extending convolution operations to meshes and graphs has been studied in both spectral-based and spatial-based fields. 
Spectral approaches build convolutional operations on spectral representations of graphs with graph Laplacian~\cite{DBLP:journals/corr/BrunaZSL13,DBLP:journals/cgf/BoscainiMMBCV15,DBLP:journals/corr/HenaffBL15,DBLP:conf/aaai/LiWZH18}.
Ranjan \emph{et al.}~\cite{DBLP:conf/eccv/RanjanBSB18} apply the truncated Chebyshev polynomials~\cite{DBLP:conf/nips/DefferrardBV16,DBLP:conf/iclr/KipfW17} to meshs.
On the other hand, spatial-based approaches take as input per vertex's local spatial structure and features to construct convolutions~\cite{DBLP:conf/nips/HamiltonYL17,DBLP:conf/iclr/VelickovicCCRLB18,DBLP:journals/corr/VermaBV17,DBLP:conf/cvpr/MontiBMRSB17,DBLP:journals/tog/WiersmaEH20,DBLP:conf/cvpr/SchultEKL20,DBLP:conf/aaai/GaoZZYYY21,DBLP:conf/nips/ZhouWLCYSLS20}.
Geodesic polar coordinates and anisotropic heat kernels are adopted to parameterize local mesh surfaces for convolutions in~\cite{DBLP:conf/iccvw/MasciBBV15,DBLP:conf/nips/BoscainiMRB16}.
Bouritsas \emph{et al.}~\cite{DBLP:conf/iccv/BouritsasBPZB19} and Gong \emph{et al.}~\cite{DBLP:conf/iccvw/GongCBZ19} propose to define local coordinates by the spiral sequence~\cite{DBLP:conf/eccv/LimDCK18}.
Convolutions on edges are defined by Hanocka \emph{et al.}~\cite{DBLP:journals/tog/HanockaHFGFC19} and extended for mesh subdivision~\cite{DBLP:journals/tog/LiuKCAJ20}.
While various convolution operators are proposed in these works, the focus of this work lies in feature aggregation across hierarchical levels, which is in parallel to them.

\par
To build hierarchical representations for meshes, Graclus algorithm~\cite{DBLP:journals/pami/DhillonGK07} and Quadric Error Metrics based mesh simplification~\cite{DBLP:conf/siggraph/GarlandH97} are adopted to reduce the number of vertices  in~\cite{DBLP:conf/nips/DefferrardBV16,DBLP:journals/corr/VermaBV17,DBLP:conf/cvpr/GeRLXWCY19,DBLP:conf/eccv/RanjanBSB18,DBLP:conf/cvpr/YuanL0DF020,DBLP:journals/corr/abs-1904-05562}.
Hanocka \emph{et al.}~\cite{DBLP:journals/tog/HanockaHFGFC19} learn to sequentially collapse edges.
These approaches aggregate features with permutation invariant functions, such as sum(·), max(·) and mean(·), for downsampling and interpolation functions for upsampling.
There are works proposing to learn the aggregation weights via a dense mapping matrix~\cite{DBLP:conf/nips/YingY0RHL18,DBLP:journals/corr/abs-1907-00481} or a fully-connected layer~\cite{DBLP:conf/cvpr/DoostiNMC20}.
As shown by our experiments, they deliver poor performance for high resolution meshes due to over-parameterization.
Zhou \emph{et al.}~\cite{DBLP:conf/nips/ZhouWLCYSLS20} directly learn aggregation weights over predefined local neighbors.
Different from these works, we learn the mapping matrices for feature aggregation through an attention based module, which is scalable due to linear complexity and 
enables learning the receptive fields together with the aggregation weights.

\section{Learning feature aggregation for 3DMMs}

\subsection{Deep 3D Morphable Models}
The deep 3D morphable model, as illustrated in Fig.~\ref{fig:hp_frame_work}, consists of an encoder and a decoder.
The encoder has a mirrored structure of the decoder. 
The encoder takes as input a 3D mesh to generate a compact representation, which captures both global and local shape information by a fine-to-coarse hierarchical network structure.
Given the compact representation, the decoder learns to generate a mesh to recover detailed shape descriptions in a coarse-to-fine fashion.
At each level of the hierarchical structure, convolution and upsampling operations are applied sequentially in the decoder, whereas convolution and downsampling operations are used for the encoder.
While the convolution can be implemented by existing isotropic~\cite{DBLP:conf/eccv/RanjanBSB18} or anisotropic~\cite{DBLP:conf/iccv/BouritsasBPZB19} convolutions, the feature aggregation, including upsampling and downsampling, is usually performed by multiplying input features by a mapping matrix.
\par
Formally, we consider a 3D mesh $\mO$ as a set of vertices and edges, $\mO=\left(\mV,\bA\right)$, where $\mV=\{1,\dots,n\}$ denotes the set of $n$ vertices and $\bA\in\left\{0,1\right\}^{n\times n}$ indicates the sparse edge connections between vertices.
Assume there are $L$ hierarchical levels in both the encoder and decoder of the morphable model.
In the following, we only take the decoder as the example to illustrate our method. However, similar analysis can be applied for the encoder.
Let $\bX^{(l)}\in\mathbb{R}^{n_{l}\times d_{l}}$ denote the output features of the decoder at level $l\in\{1,2,\dots,L\}$, where $n_{l}$ and $d_{l}$ are the number of vertices and the feature dimension.
$\bX^{(l-1)}$ serves as the input features at level $l$. 

\par
In existing models~\cite{DBLP:conf/eccv/RanjanBSB18,DBLP:conf/iccv/BouritsasBPZB19},  feature aggregation can be generally formulated as
\begin{equation}\label{eqn_general_aggr}
    \bx^{(l)}_i = \sum_{j=1}^{n_{l-1}} m_{ij}^{(l-1\rightarrow l)} \bx^{(l-1)}_{j},
\end{equation}
with different aggregation weights $m^{(l-1\rightarrow l)}_{ij}\in\bM^{(l-1\rightarrow l)}$ assigned for downsampling and upsampling.
While these models demonstrate promising performance, the mapping matrices $\bM^{(l-1\rightarrow l)}$ are calculated by mesh simplification on a template mesh and are fixed during the training of the model.
In contrast, we are interested in learning these mapping matrices along with the convolution operations for the training shapes.
A straightforward solution would be directly parameterizing the mapping matrices as in~\cite{DBLP:conf/nips/YingY0RHL18,DBLP:conf/cvpr/DoostiNMC20} and training them along with the other components of the model.
However, this suffers from over-parameterization for high resolution meshes and deteriorates the model performance as we show in the experiments.
On the contrary, we propose to learn the mapping matrices for feature aggregation through an attention based module.

\subsection{Learning feature aggregation via attention}
\begin{figure}[tb]
    \centering
    \includegraphics[width=.95\linewidth]{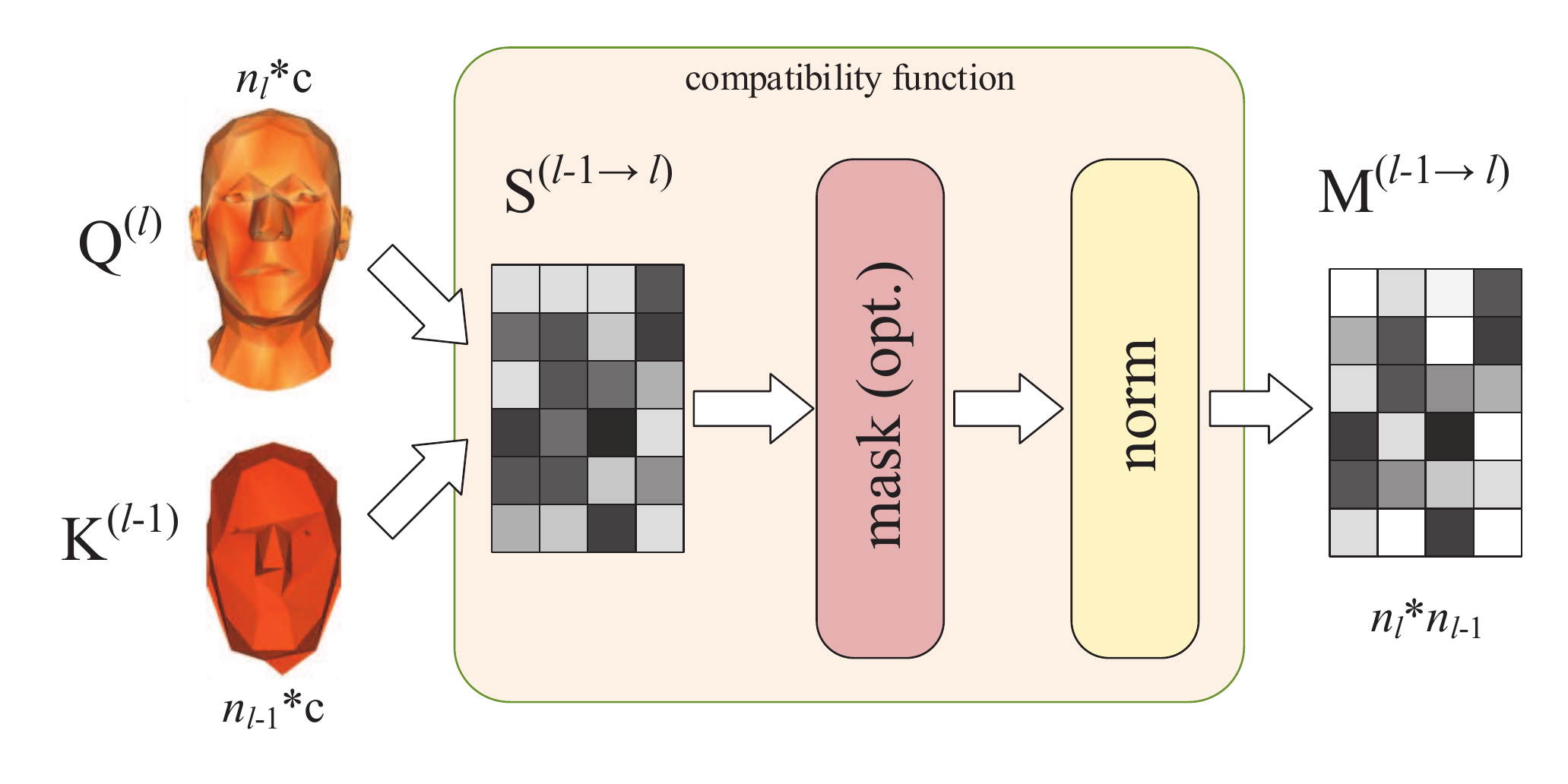}
    \caption{The attention based feature aggregation module}
    \label{fig:hp_module}
\end{figure}

We propose to model the feature aggregation from $\bX^{(l-1)}$ to $\bX^{(l)}$ by the attention mechanism~\cite{DBLP:conf/nips/SukhbaatarSWF15,DBLP:conf/nips/VaswaniSPUJGKP17,DBLP:conf/iclr/VelickovicCCRLB18}.
The attention mechanism can be regarded as generating an output with a corresponding query vector and the stored key-value pairs.
In the context of feature aggregation, $\bx_i^{(l)}$ is the output and $\bx_{j}^{(l-1)}, j=1,\dots,n_{l-1}$ are the stored values. 
Then, we propose to compute the aggregation weight $m_{ij}^{(l-1\rightarrow l)}$ by a compatibility function $d(\cdot)$ of a query vector $\bq_i^{(l)}\in\mathbb{R}^{c}$ and the key vector $\bk_j^{(l-1)}\in\mathbb{R}^{c}$ that corresponds to the stored value $\bx_{j}^{(l-1)}$.
The formulation is as
\begin{equation}\label{eqn_att_dist}
    m_{ij}^{(l-1\rightarrow l)} = d\left(\bq_i^{(l)}, \bk_j^{(l-1)}\right).
\end{equation}
By this means, we can break down the construction of mapping matrices into the establishment of the key and query vectors and the design of the compatibility function, as shown in Fig.~\ref{fig:hp_module}.

\subsubsection{The key and query vectors}

Since the key and query vectors are paired with the vertex features, their values are expected to be either associated with the vertex indices or related to the vertex features.
Given the fact that feature irrelevant mapping matrices are feasible in the existing models~\cite{DBLP:conf/eccv/RanjanBSB18,DBLP:conf/iccv/BouritsasBPZB19}, we model the key and query vectors as a function of the vertex indices rather than the vertex features. 
In other words, we aim to learn object aware mapping matrices instead of instance aware ones.
Recall that the meshes considered in this work are already registered, i.e., each vertex has its corresponding semantic meaning.
The key and query vectors can, then, be parameterized as trainable variables without taking any data as input.
One benefit of such parameterization is that only the learned mapping matrices rather than the attention module are required at inference stage. 
Furthermore, this also avoids the chicken-and-egg situation between $\bq_i^{(l)}$ and $\bx_i^{(l)}$ when associating queries with vertex features.
While the numbers of key and query vectors are determined by the numbers of vertices, the feature dimension of each vector $c$ is a hyperparameter to be set.

\subsubsection{The compatibility function}
Given the key and query vectors, the compatibility function measures how well two vertices at neighboring levels align.
In general, arbitrary distance metric can be employed to compute the compatibility score.
We propose to compute the compatibility score with the cosine distance between $\bq_i^{(l)}\in\mathbb{R}^{c}$ and $\bk_j^{(l-1)}\in\mathbb{R}^{c}$ as
\begin{equation}\label{eqn_weight}
    s^{(l-1\rightarrow l)}_{w,ij} = \cos\left(\bq_i^{(l)} , \bk_j^{(l-1)}\right).
\end{equation}
Directly applying this score as the aggregation weight would result in a dense mapping matrix. 
In other words, any vertex feature in a given level is influenced by all vertex features in the preceding level.
However, such aggregation is less desirable in the context of hierarchical learning.
\par
In order to capture both the global and local information, the receptive fields of different $\bx_i^{(l)}$ are expected to be distinctive to each other.
Each vertex in the succeeding level $\bx_i^{(l)}$ is expected to be related to an unique subset of the vertices in the preceding level.
We integrate such prior knowledge into an optional mask operation by setting the cardinality of the subset to a fixed value of $k$ with top-$k$ selection.
And the elements of each subset are automatically learned from the training shapes.
To be specific, we define a binary mask for the weight score $s^{(l-1\rightarrow l)}_{w,ij}$ as
\begin{equation}\label{eqn_pool_mask_func}
    b_{ij}^{(l-1\rightarrow l)} = \begin{cases}
    1, & \text{$s^{(l-1\rightarrow l)}_{w,ij}$ is among the top $k$ of $s^{(l-1\rightarrow l)}_{w,i:}$}\\
    0, &\text{otherwise}.
    \end{cases}
\end{equation}
The masked weight score is then computed by multiplying the weight score by the binary mask as
\begin{equation}\label{eqn_pool_masked}
    s_{m,ij}^{(l-1\rightarrow l)} = b_{ij}^{(l-1\rightarrow l)} s_{w,ij}^{(l-1\rightarrow l)}.
\end{equation}
We then normalize the weight score as
\begin{equation}\label{eqn_attention_mapping}
    m^{(l-1\rightarrow l)}_{a,ij} = \frac{s^{(l-1\rightarrow l)}_{m,ij}}{\sum_{j=1}^{n_{l-1}}s^{(l-1\rightarrow l)}_{m,ij}}.
\end{equation}

\subsubsection{The mapping matrix}
While it is feasible to take Eqn.~\ref{eqn_attention_mapping} as the replacement of the precomputed weights in existing models, we further propose to fuse both information to leverage the success of existing methods. The fusion can be thought as a multi-head attention with a fixed head and a learnable head.
Specifically, we adopt a weighted combination for fusion as
\begin{equation}\label{eqn_weighted_mapping}
    m_{ij}^{(l-1\rightarrow l)} = w_a m_{a,ij}^{(l-1\rightarrow l)} + (1-w_a) m_{p,ij}^{(l-1\rightarrow l)},
\end{equation}
where $w_a$ is a trainable weight and $m_{p,ij}^{(l-1\rightarrow l)}$ stands for the precomputed downsampling or upsampling coefficient in existing models~\cite{DBLP:conf/iccv/BouritsasBPZB19,DBLP:conf/eccv/RanjanBSB18}.
When $w_a$ is a fixed value of 0, Eqn.~\ref{eqn_weighted_mapping} is exactly the same as the precomputed ones in existing models.
With the aggregation weight computed according to Eqn.~\ref{eqn_weighted_mapping}, we can obtain the mapping matrices $\bM^{(l-1\rightarrow l)}$ for both upsampling and downsampling in the model.

\section{Experiments}
In this section, we demonstrate the effectiveness of our proposed attention based feature aggregation for 3D reconstruction on human face, human body, and hand shape datasets in comparison to existing 3DMMs~\cite{DBLP:conf/eccv/RanjanBSB18,DBLP:conf/iccv/BouritsasBPZB19}.

\begin{figure*}[tb]
\centering
{
    \label{fig_rms_coma}
    \begin{minipage}[t]{5.4cm}
    \centering
    \includegraphics[width=5.4cm]{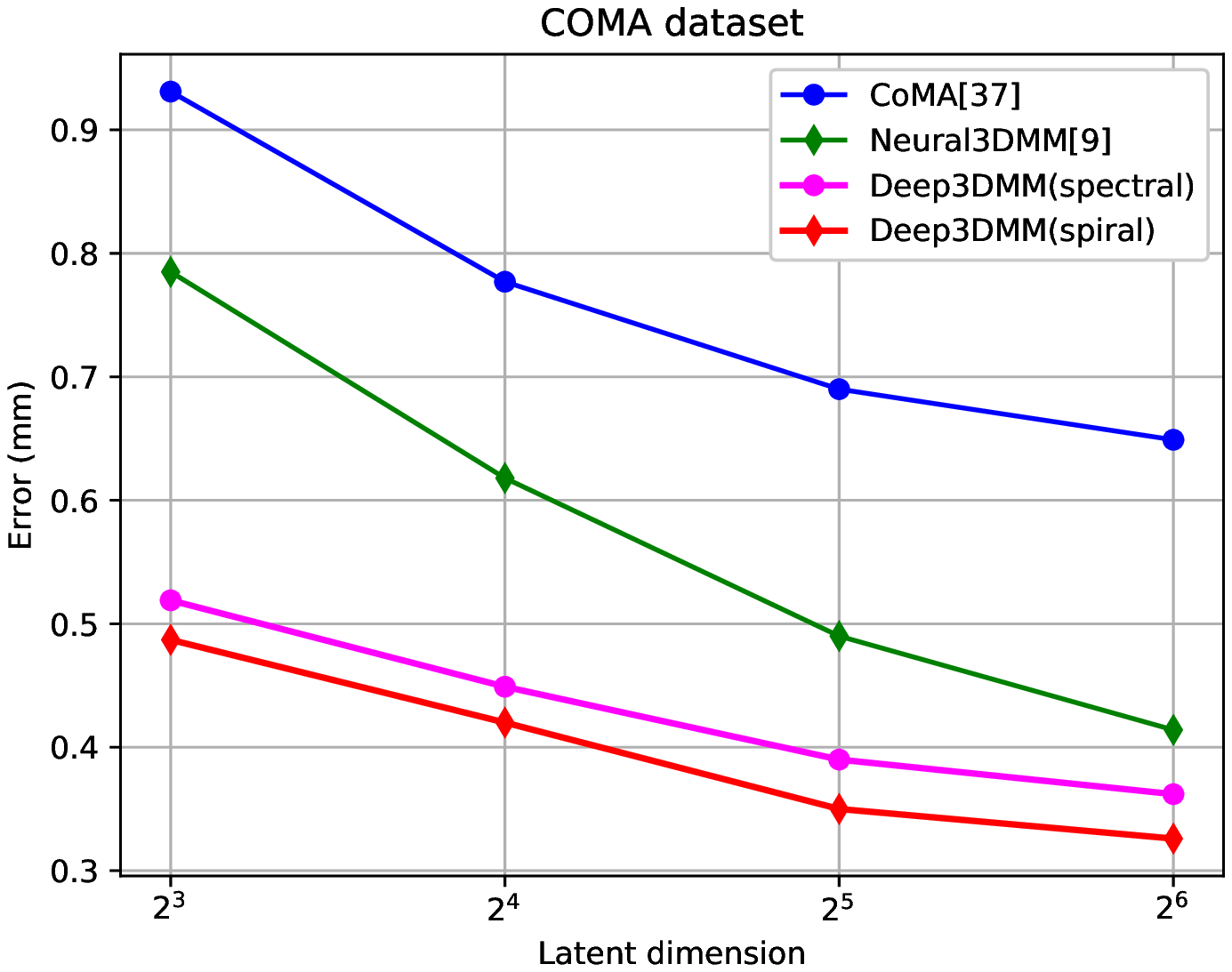}
    \end{minipage}
}
\hspace{0.1cm}
{
    \label{fig_rms_dfaust}
    \begin{minipage}[t]{5.4cm}
    \includegraphics[width=5.4cm]{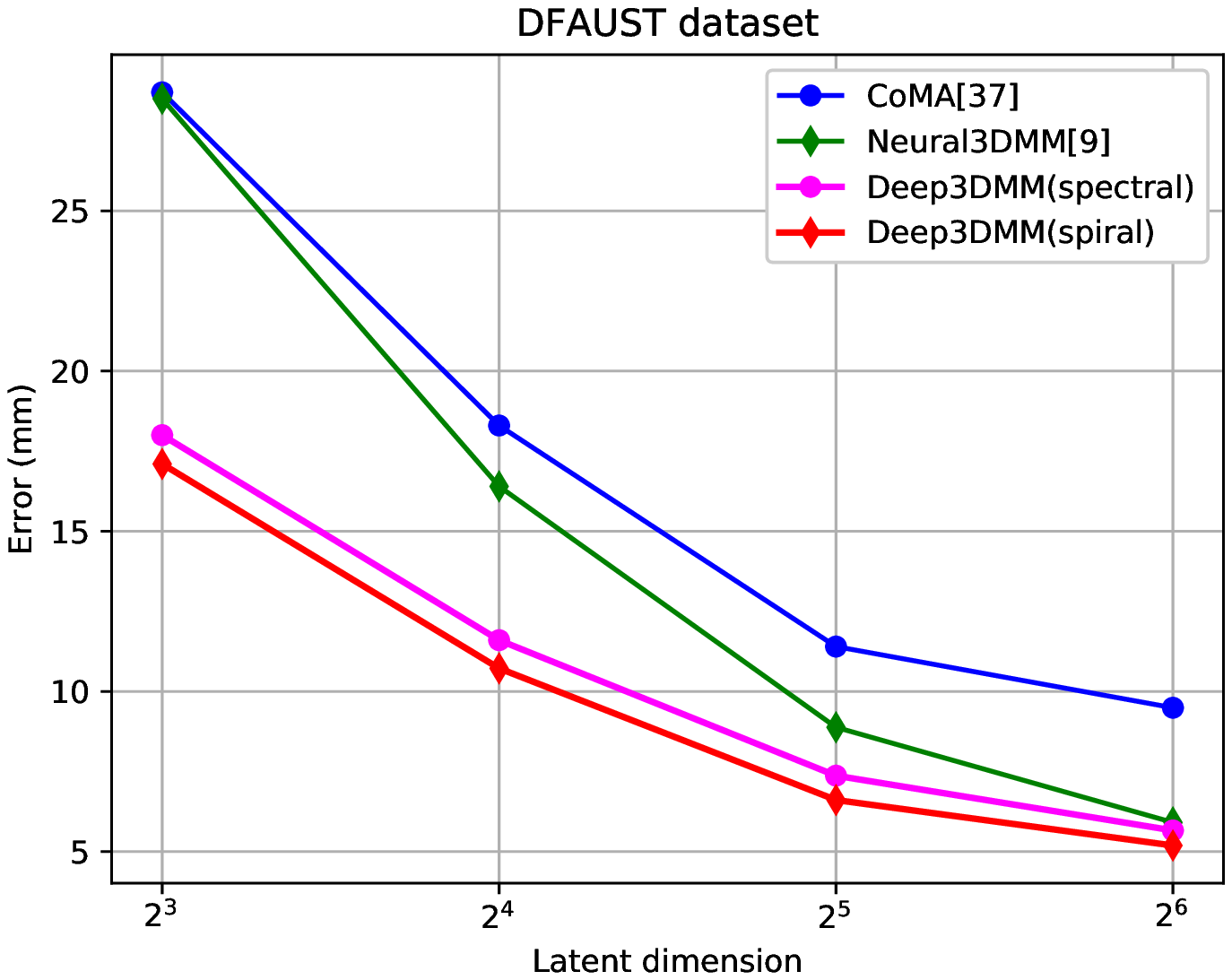}
    \end{minipage}
}
\hspace{0.1cm}
{
    \label{fig_rms_synhand}
    \begin{minipage}[t]{5.4cm}
    \includegraphics[width=5.4cm]{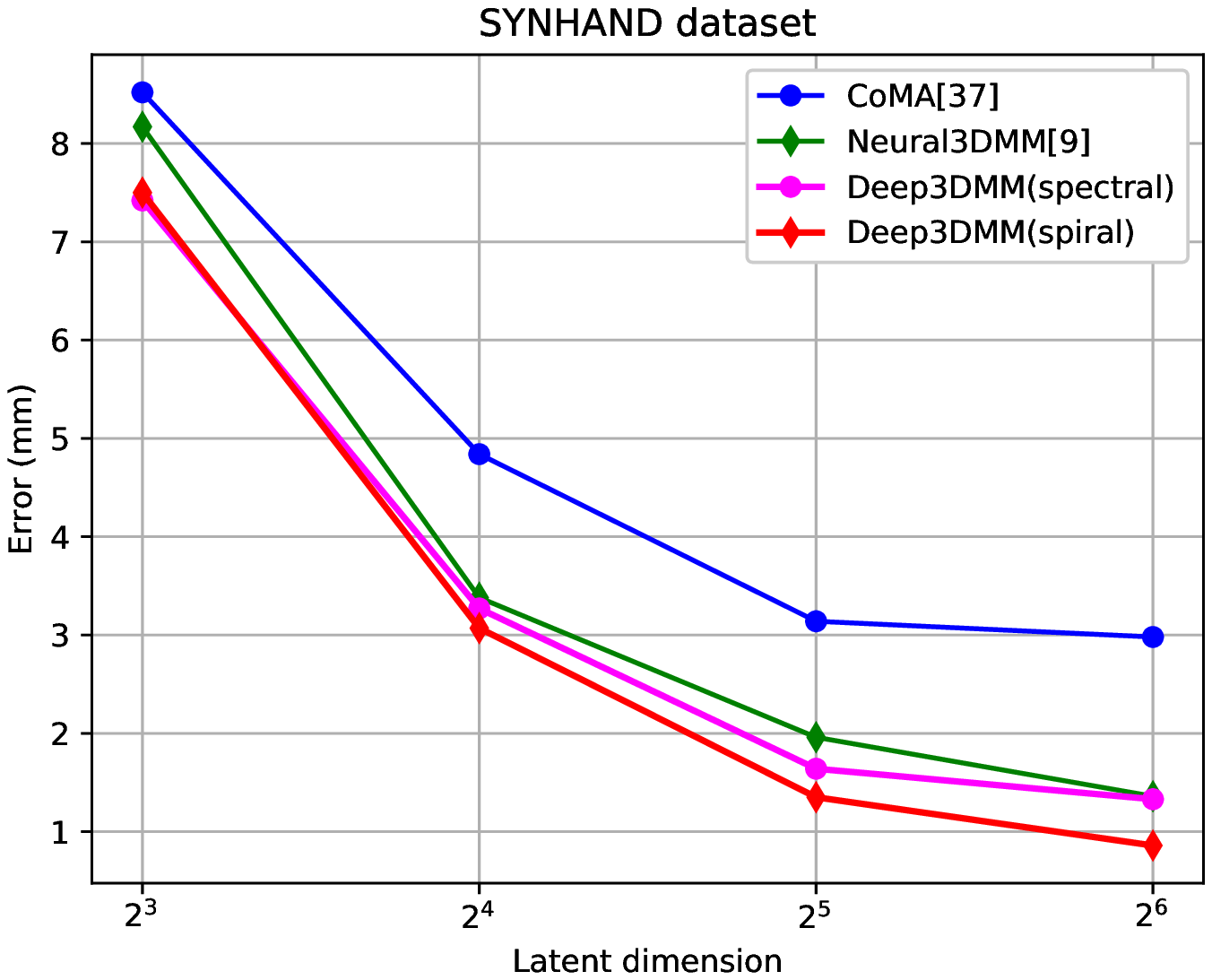}
    \end{minipage}
}
\caption{Reconstruction errors of different methods on COMA (left), DFAUST (middle), and SYNHAND (right) datasets}
\label{fig_rms}
\end{figure*}

\subsection{Datasets}
\textbf{COMA~\cite{DBLP:conf/eccv/RanjanBSB18}.} The face expression dataset presented by Ranjan \emph{et al.}~\cite{DBLP:conf/eccv/RanjanBSB18} consists of 20K+ well registered 3D face shapes with 5,023 vertices for each shape.
The dataset provides twelve extreme expression sequences, such as high smile and bareteeth, from twelve unique identities.
We follow the interpolation setting in~\cite{DBLP:conf/eccv/RanjanBSB18} to split the dataset.
\par
\textbf{DFAUST~\cite{DBLP:conf/cvpr/Bogo0PB17}.} This dataset contains 40K+ dynamic human body shapes with all shapes aligned to a common reference topology with 6,890 vertices~\cite{DBLP:conf/cvpr/Bogo0PB17}.
The shapes are captured from ten identities in over one hundred moving sequences, covering actions like punching and shaking shoulders and hips.
We follow the data split setting in~\cite{DBLP:conf/iccv/BouritsasBPZB19}.
\par
\textbf{SYNHAND~\cite{DBLP:conf/3dim/MalikENVTHS18}.} A total of 5 million synthetic hand meshes with varying hand shapes and poses are presented in~\cite{DBLP:conf/3dim/MalikENVTHS18}. We follow~\cite{DBLP:conf/eccv/TretschkTZGT20} to randomly sample 100K meshes for the experiments with hands, where 90K meshes are used for training and validation and the remaining 10K meshes are used to evaluate the trained model.

\subsection{Implementation details}
We choose to evaluate the proposed feature aggregation method on two state-of-the-art 3D morphable models with isotropic spectral convolution~\cite{DBLP:conf/eccv/RanjanBSB18} and anisotropic spiral convolution~\cite{DBLP:conf/iccv/BouritsasBPZB19}, denoted as Deep3DMM(spectral) and Deep3DMM(spiral), respectively.
We follow the settings in~\cite{DBLP:conf/eccv/RanjanBSB18} to configure the network architecture as an encoder-decoder with four downsampling and upsampling layers.
The mapping matrices are generated by our method while the
remaining components of the model are consistent with existing models.
The whole network is trained with the recommended settings in the original works~\cite{DBLP:conf/eccv/RanjanBSB18,DBLP:conf/iccv/BouritsasBPZB19}.
\par
For our attention based feature aggregation module, we empirically set $c$ to 21 and $k$ to 2 and 32 for downsampling and upsampling, respectively.
The trainable weight $w_a$ is initialized as $0.2$.
Different from the standard initialization of the remaining components in the network, we use the results of mesh simplification to initialize the key and query vectors.
In particular, we initialize the first three dimensions of the key and query vectors by the spatial position of vertices at the corresponding level. The remaining dimensions are randomly initialized by an uniform distribution.

\subsection{Results of attention based feature aggregation}
\subsubsection{Quantitative results of reconstruction}
We follow previous works~\cite{DBLP:conf/eccv/RanjanBSB18,DBLP:conf/iccv/BouritsasBPZB19} to evaluate the performance of shape representation in terms of Euclidean distance based reconstruction error on the test set.
In Fig.~\ref{fig_rms}, we show the quantitative results of the averaged reconstruction error on three datasets with the dimension of latent representation varying from 8 to 64.
\par
As shown in the figure, morphable models equipped with our attention based feature aggregation module consistently outperform the corresponding baseline models under all evaluated settings.
For the spectral convolution based models, our feature aggregation reduces the reconstruction errors by a large margin for all tested latent dimensions on all three datasets (60\% lower on COMA dataset).
That should be attributed to the mechanism of learning both the receptive fields and the aggregation weights.
This mechanism allows each vertex to be affiliated with a selected portion of vertices at the preceding level with different weights, which compensates the isotropy of spectral convolutions.
For the spiral convolution based models, our feature aggregation also enhances the model capacity in every tested case.
A possible reason of the improvement is the adaptive learning of the receptive field for each vertex.
\par
We also observe that the gap between spectral convolution based model and spiral convolution based model is narrowed after applying our feature aggregation.
We attribute this to the fact that our feature aggregation also introduces certain anisotropic operations which partly overlap with the spiral convolutions.
Moreover, we observe that the reduced error introduced by our feature aggregation, especially in combination with spiral convolutions, is more significant for shorter latent representations, which may benefit downstream tasks to capture semantically meaningful and discriminative representations with a lightweight model.

\begin{figure}[tb]
\centering
\includegraphics[width=.98\linewidth]{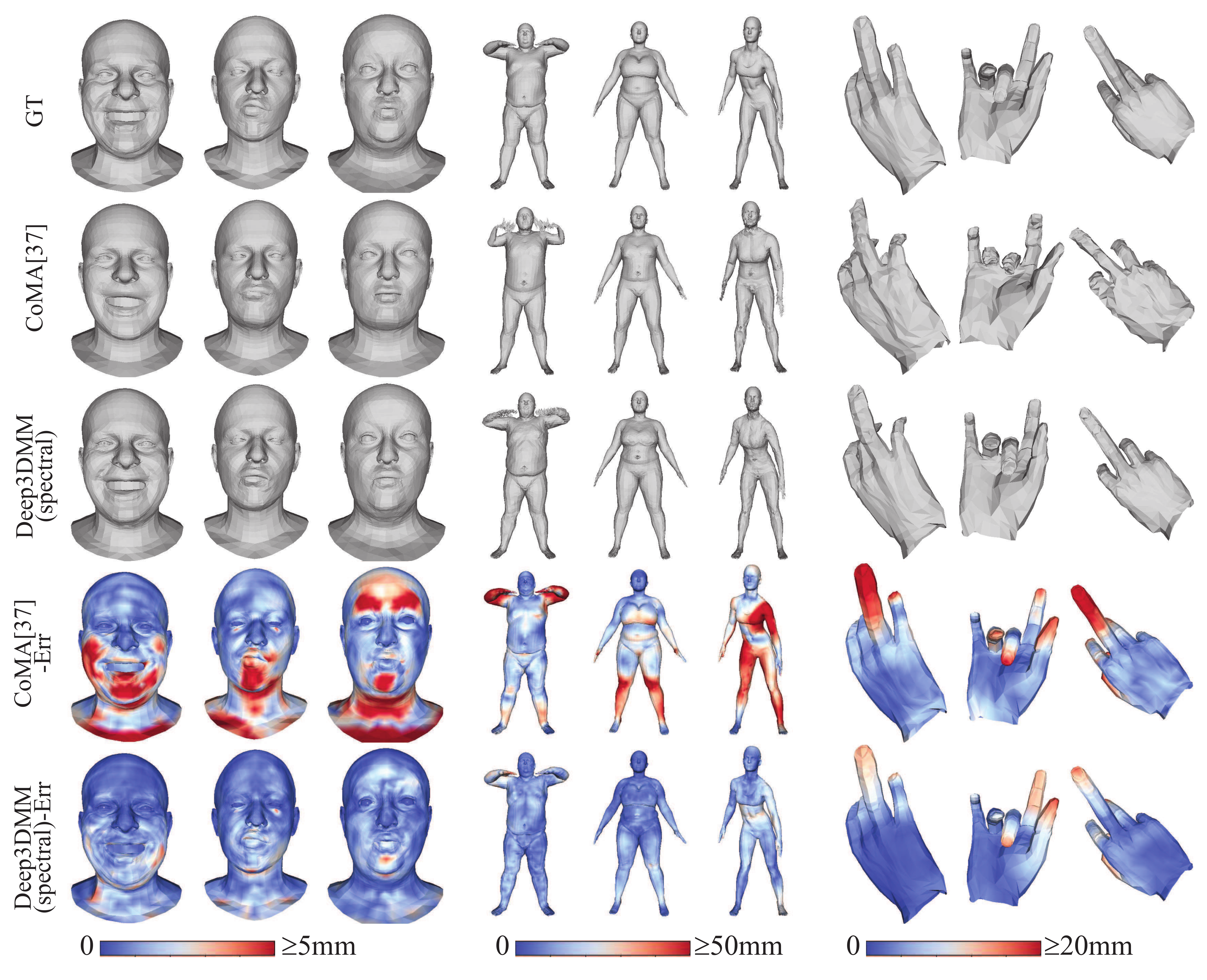}
\caption{Qualitative comparison of spectral convolution based models on COMA (left), DFAUST (middle), and SYNHAND (right) datasets. The first row is the ground truth shapes. The second and third rows show the reconstructed shapes, while the fourth and fifth rows show the corresponding reconstruction errors}
\label{fig_viz_recon_spectral}
\end{figure}

\subsubsection{Qualitative results of reconstruction}%
In Fig.~\ref{fig_viz_recon_spectral},
we visualize the per vertex Euclidean error of different morphable models on several shapes from the three datasets for qualitative comparison. The latent dimension is set as 8.
We can observe that our feature aggregation reduces the reconstruction error not only on the vertices with large errors but also on other vertices.
While existing morphable models may have wrong predictions in some parts of the shape, such as the forehead of face, the hands and legs of human, and the fingers of hand, our proposed feature aggregation can alleviate this problem by providing rough correct predictions.
Meanwhile, our feature aggregation leads to more realistic reconstructions by recovering more details.

\subsubsection{Visualization of mapping matrices}

\begin{figure}[tb]
\centering
\includegraphics[width=.9\linewidth]{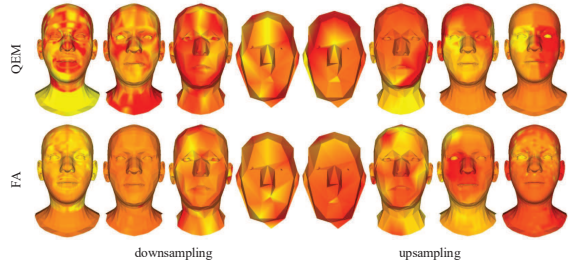}
\caption{Visualization of mapping matrices of down-sampling and up-sampling on COMA dataset with t-SNE. The first and second rows are the results of quadric error minimization in~\cite{DBLP:conf/eccv/RanjanBSB18,DBLP:conf/iccv/BouritsasBPZB19} and our proposed feature aggregation, respectively }
\label{fig_viz_mapping_hierarchy_coma}
\end{figure}

\begin{figure*}[tb]
\centering
\includegraphics[width=.95\linewidth]{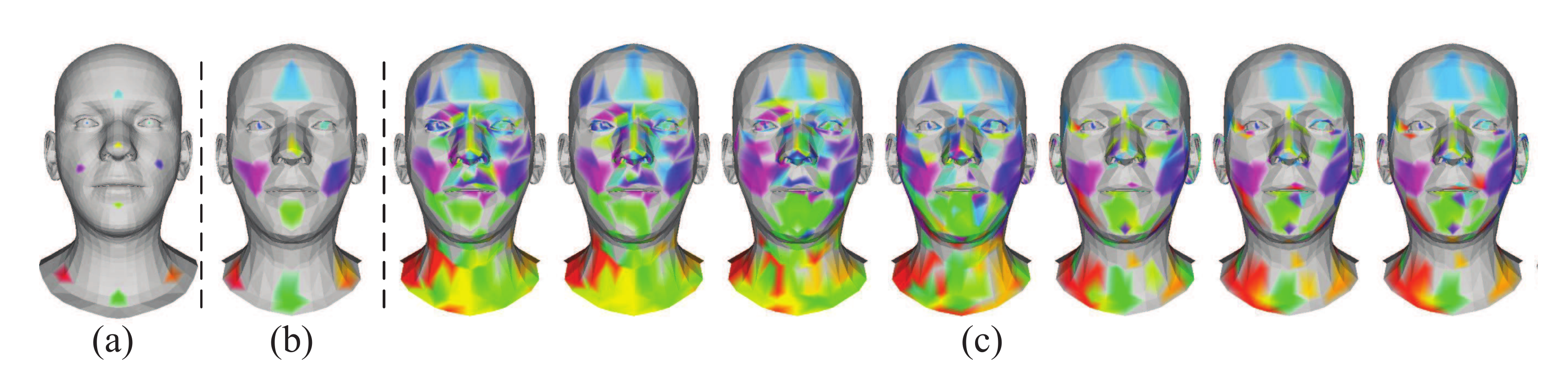}
\caption{Illustrations of receptive fields of selected vertices at the finest level of upsampling on COMA dataset (best viewed in color). The first column (a) shows the selected vertices with colors. The second column (b) shows the receptive fields of the existing QEM method at the neighboring coarser level. The remaining columns (c) show the receptive fields of our proposed attention based feature aggregation at epochs of 1, 50, 100, 150, 200, 250, and 300 (from left to right) with convolutions in~\cite{DBLP:conf/eccv/RanjanBSB18}}
\label{fig_viz_mapping_coma}
\end{figure*}

To understand the learning of mapping matrices for feature aggregation, we provide the visualization of mapping matrices in Figs.~\ref{fig_viz_mapping_hierarchy_coma} and~\ref{fig_viz_mapping_coma} with the spectral convolution on COMA dataset.
In Fig.~\ref{fig_viz_mapping_hierarchy_coma}, we show the mapping matrices of both downsampling and upsampling on the corresponding shapes for both quadric error minimization (QEM) in existing morphable models~\cite{DBLP:conf/eccv/RanjanBSB18,DBLP:conf/iccv/BouritsasBPZB19} and our proposed feature aggregation (FA).
We map the column/row vectors of the mapping matrices of down-/up-sampling by t-SNE to a 1-dimension manifold for visualization.
We can observe that the learned mapping matrices are notably different from those computed by mesh decimation.
The learned aggregation weights are more likely to be similar for vertices that are close to each other, such as the region around eyes in the rightmost shape.

\par
In Fig.~\ref{fig_viz_mapping_coma}, we visualize the receptive fields of some exemplar vertices for the last level of upsampling.
Since the traced downsampling and barycentric coordinates are used for upsampling in QEM, we can find that the receptive fields are small local regions around the traced vertices, as shown in Fig.~\ref{fig_viz_mapping_coma}(b).
In contrast, we can observe that the receptive fields are randomly distributed on the face with random initialization of the keys and queries in our attention based feature aggregation, as shown in the leftmost of Fig.~\ref{fig_viz_mapping_coma}(c).
By training the feature aggregation together with the convolutions, most of the receptive fields also converge to the local regions around the traced vertices, but with larger sizes than that of QEM.
Besides the geodesic neighboring vertices, there are some non-local receptive fields, such as vertices at the back of head as shown in the bottom row.
Note that the training of mapping matrices is based on the vertex features, rather than the spatial information of each vertex.
It is possible that non-local vertices contain the identity and expression information that is helpful for recovering shape details.
Along with the training of receptive fields, the aggregation weights are simultaneously trained for the receptive fields.
This is shown in the figure by the variation of saturation and brightness, for example on the chin and neck.

\subsection{Comparison with other aggregation methods}
In order to show the effectiveness of simultaneously learning receptive fields and aggregation weights by the attention mechanism, we also conduct experiments with different aggregation methods.
In Table~\ref{table_aggr}, we show the results with latent dimension of 8 on COMA dataset with spectral convolutions.
Full mapping method directly parameterizes the mapping matrices as trainable parameters to learn both the aggregation weights and receptive fields.
However, this method delivers poor performance since it suffers from over-parameterization.
We migrate DiffPool~\cite{DBLP:conf/nips/ZhouWLCYSLS20} for mesh autoencoders by learning the mapping matrices with additional network layers as suggested by the authors. This method also does not perform well due to learning the matrics directly.
Average aggregation uses the same receptive fields as QEM, but the assigned aggregation weights are identical for vertices in the receptive field.
QEM in~\cite{DBLP:conf/eccv/RanjanBSB18,DBLP:conf/iccv/BouritsasBPZB19} adopts mesh decimation for downsampling and barycentric coordinates for upsampling.
Variant weight~\cite{DBLP:conf/nips/ZhouWLCYSLS20} automatically learns the aggregation weights by setting them as trainable parameters. Since the local region defined in~\cite{DBLP:conf/nips/ZhouWLCYSLS20} relies on extra connection matrices between hierarchical levels which are unavailable in existing models~\cite{DBLP:conf/eccv/RanjanBSB18,DBLP:conf/iccv/BouritsasBPZB19}, we set 1-ring neighbors as the local region.
The main drawback of these methods is the inability to learn the receptive fields.
We do not compare with MeshCNN~\cite{DBLP:journals/tog/HanockaHFGFC19} since it is unclear how to incorporate the edge based operations to existing models~\cite{DBLP:conf/eccv/RanjanBSB18,DBLP:conf/iccv/BouritsasBPZB19} defined on vertices. It is also computation inefficient and worse than Neural3DMM~\cite{DBLP:conf/iccv/BouritsasBPZB19} as show in~\cite{DBLP:conf/nips/ZhouWLCYSLS20}.
Our proposed feature aggregation balances well between the number of parameters and the flexibility to learn receptive fields and aggregation weights via the attention mechanism.

\setlength{\tabcolsep}{4pt}
\begin{table}
\begin{center}
\caption{Comparison of different aggregation methods. $\ast$ denotes our implementation of the method for the mesh autoencoder in this paper}
\label{table_aggr}
\begin{tabular}{lr}
\toprule[1pt]
Aggregation method    & Error (mm)    \\
\midrule[0.5pt]
Full mapping        & 4.319   \\
DiffPool~\cite{DBLP:conf/nips/YingY0RHL18}$^\ast$        & 3.964   \\
Average    & 0.975  \\
QEM~\cite{DBLP:conf/eccv/RanjanBSB18}    & 0.931   \\
Variant weight~\cite{DBLP:conf/nips/ZhouWLCYSLS20}$^\ast$        & 0.928   \\
Ours     & \textbf{0.519}  \\
\bottomrule[1pt]
\end{tabular}
\end{center}
\end{table}

\subsection{Results with different settings of filters}
We also conduct experiments on two settings of the number of convolution filters to explore the effectiveness of our proposed feature aggregation mechanism with different networks.
The simple setting represents the configuration as in~\cite{DBLP:conf/eccv/RanjanBSB18}, where the number of filters are (3,16,16,16,32) and (32,32,16,16,16,3) for the encoder and decoder, respectively.
The wider setting represents the network with larger numbers of filters, where they are (3,16,32,64,128) and (128,64,32,32,16,3) for the encoder and decoder, respectively.
Table~\ref{table_filter_spec} shows the reconstruction errors on COMA dataset with the latent dimension of 8, together with the number of parameters at the inference stage.
We can see that our feature aggregation mechanism consistently reduces the error under both settings with the same number of parameters at the inference stage.

\setlength{\tabcolsep}{4pt}
\begin{table}
\begin{center}
\caption{Reconstruction errors with different settings of filters }
\label{table_filter_spec}
\begin{tabular}{l|cc|cc}
\toprule[1pt]
      & \multicolumn{2}{c|}{simple}    & \multicolumn{2}{c}{wider}    \\
Method & error & \# of par.  & error & \# of par.              \\
\midrule[0.5pt]
CoMA~\cite{DBLP:conf/eccv/RanjanBSB18}      & 0.939 &30,056 & 0.682 & 179,656         \\ 
Ours   & 0.519 &30,056     & 0.441 & 179,656  \\
\bottomrule[1pt]
\end{tabular}
\end{center}
\end{table}

\setlength{\tabcolsep}{4pt}
\begin{table}
\begin{center}
\caption{Ablation study of the feature aggregation }
\label{table_coma_ablation_att}
\begin{tabular}{l|cccc}
\toprule[1pt]
      & \multicolumn{4}{c}{encoder/decoder} \\
Method & QEM/QEM & FA/QEM & QEM/FA & FA/FA              \\
\midrule[0.5pt]
Error (mm)   & 0.939 &0.886     & 0.525 & 0.519 \\

\bottomrule[1pt]
\end{tabular}
\end{center}
\end{table}

\subsection{Ablation study of the feature aggregation}
In Table~\ref{table_coma_ablation_att}, we show the reconstruction errors by applying the proposed feature aggregation (FA) on either the encoder or the decoder.
The results show that the performance can be enhanced by learning the feature aggregation on either the encoder or the decoder, especially on the decoder.

\subsection{Latent space arithmetic}
We also evaluate the representational power of the learned latent space by performing arithmetic operations on latent representations.

\begin{figure}[tb]
\centering
\includegraphics[width=.9\linewidth]{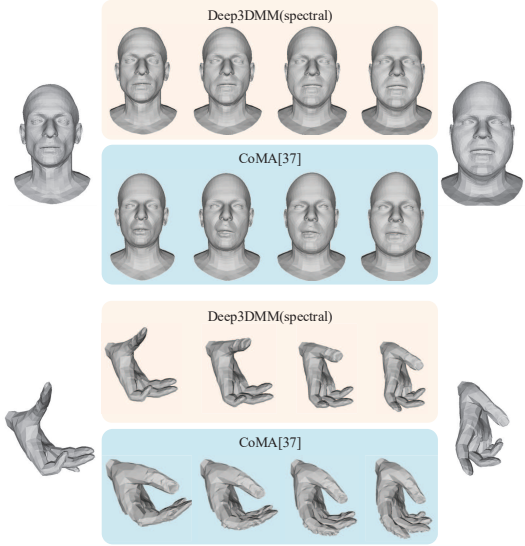}
\caption{Interpolations between different shapes. The left-most and right-most shapes are input shapes}
\label{fig_viz_interpolations}
\end{figure}

\noindent\textbf{Interpolation.} Given the latent representations $\bz_1$ and $\bz_2$ of two sufficient different shapes $\mO_1$ and $\mO_2$, we linearly interpolate between them to obtain a new latent representation $\bz_0=\alpha \bz_1+(1-\alpha)\bz_2$ with $\alpha\in(0,1)$. As shown in Fig.~\ref{fig_viz_interpolations}, by decoding $\bz_0$ with the mesh decoder, we can obtain the in-between shapes for different $\alpha$. The results of our model are better.

\begin{figure}[tb]
\centering
\includegraphics[width=.9\linewidth]{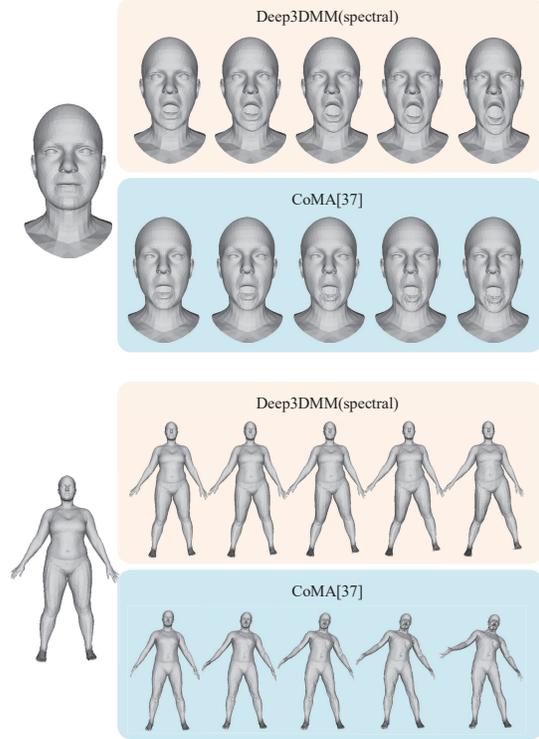}
\caption{Extrapolation. The left-most shapes are the neutral shapes}
\label{fig_viz_extrapolations}
\end{figure}

\noindent\textbf{Extrapolation.} Similar to the linear combination for interpolation, we obtain an extrapolated latent representation by performing  $\bz_0=\alpha \bz_1+(1-\alpha)\bz_2$ but with $\alpha\in(-\infty,0)\cup(1,\infty)$. As shown in Fig.~\ref{fig_viz_extrapolations}, we choose $\bz_1$ to be the neutral shape of the same identity as $\bz_2$ and show the overdrawn shape deformation of a given identity. The results of our model are more natural.

\begin{figure}[tb]
\centering
\includegraphics[width=.9\linewidth]{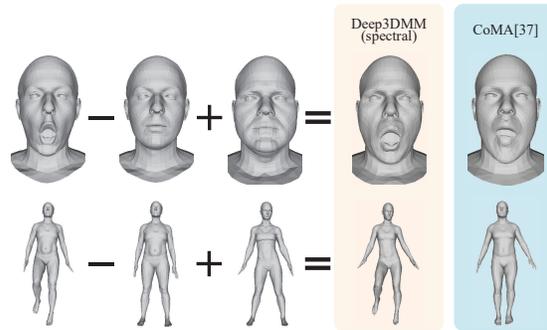}
\caption{Deformation transfer with arithmetic operations on latent representations}
\label{fig_viz_analog}
\end{figure}

\noindent\textbf{Deformation transfer.} It is also possible to transfer shape deformation between different identities. Let $\mS_0,\mT_0$ be the shapes of two different identities with the neutral expression/pose, $\mS_1$ be the shape of the same identity as $\mS_0$ but with a specific expression/pose.
We compute the shape deformation by $d=\mS_1-\mS_0$ and apply this deformation to $\mT_0$ by $\mT_1=\mT_0+d$ to get the deformed shape.
Examples of face and human body are shown in Fig.~\ref{fig_viz_analog}. Our model can successfully transfer the deformation while the baseline model fails.

\section{Conclusion}

In this paper, we propose to learn feature aggregation for deep 3D morphable models.
Specifically, we introduce keys and queries as trainable parameters and adopt the attention mechanism to compute the mapping matrices for vertex feature upsampling and downsampling.
Our attention based feature aggregation allows the receptive fields and aggregation weights to be simultaneously learned.
Our experiments on three benchmarks show that the proposed feature aggregation can enhance the capacity of 3D morphable models with either isotropic or anisotropic convolutions.

There are several possible future directions for our work.
While the attention mechanism is introduced for feature aggregation in this work, it would be interesting to apply the attention mechanism for the other components of the morphable models.
In addition, while current models require the input shapes to be registered to a template mesh, we are also interested in extending our method to arbitrary meshes.

\section*{Acknowledgment}
This work was supported by a Huawei research grant.

{\small
\bibliographystyle{ieee_fullname}
\bibliography{egbib}

\begin{thebibliography}{10}\itemsep=-1pt

\bibitem{DBLP:conf/fgr/AmbergKV08}
Brian Amberg, Reinhard Knothe, and Thomas Vetter.
\newblock Expression invariant {3D} face recognition with a morphable model.
\newblock In {\em {FG}}, pages 1--6, 2008.

\bibitem{DBLP:journals/corr/abs-1907-00481}
Filippo~Maria Bianchi, Daniele Grattarola, and Cesare Alippi.
\newblock Mincut pooling in graph neural networks.
\newblock {\em CoRR}, abs/1907.00481, 2019.

\bibitem{DBLP:conf/siggraph/BlanzV99}
Volker Blanz and Thomas Vetter.
\newblock A morphable model for the synthesis of {3D} faces.
\newblock In {\em {SIGGRAPH}}, pages 187--194, 1999.

\bibitem{DBLP:conf/cvpr/Bogo0PB17}
Federica Bogo, Javier Romero, Gerard Pons{-}Moll, and Michael~J. Black.
\newblock Dynamic {FAUST:} registering human bodies in motion.
\newblock In {\em CVPR}, pages 5573--5582, 2017.

\bibitem{DBLP:journals/ijcv/BoothRPDZ18}
James Booth, Anastasios Roussos, Allan Ponniah, David Dunaway, and Stefanos
  Zafeiriou.
\newblock Large scale {3D} morphable models.
\newblock {\em IJCV}, 126(2-4):233--254, 2018.

\bibitem{DBLP:journals/cgf/BoscainiMMBCV15}
Davide Boscaini, Jonathan Masci, Simone Melzi, Michael~M. Bronstein, Umberto
  Castellani, and Pierre Vandergheynst.
\newblock Learning class-specific descriptors for deformable shapes using
  localized spectral convolutional networks.
\newblock {\em Comput. Graph. Forum}, 34(5):13--23, 2015.

\bibitem{DBLP:conf/nips/BoscainiMRB16}
Davide Boscaini, Jonathan Masci, Emanuele Rodol{\`{a}}, and Michael~M.
  Bronstein.
\newblock Learning shape correspondence with anisotropic convolutional neural
  networks.
\newblock In {\em NeurIPS}, pages 3189--3197, 2016.

\bibitem{DBLP:journals/tog/BouazizWP13}
Sofien Bouaziz, Yangang Wang, and Mark Pauly.
\newblock Online modeling for realtime facial animation.
\newblock {\em TOG}, 32(4):40:1--40:10, 2013.

\bibitem{DBLP:conf/iccv/BouritsasBPZB19}
Giorgos Bouritsas, Sergiy Bokhnyak, Stylianos Ploumpis, Stefanos Zafeiriou, and
  Michael~M. Bronstein.
\newblock Neural {3D} morphable models: Spiral convolutional networks for {3D}
  shape representation learning and generation.
\newblock In {\em {ICCV}}, pages 7212--7221, 2019.

\bibitem{DBLP:journals/corr/BrunaZSL13}
Joan Bruna, Wojciech Zaremba, Arthur Szlam, and Yann LeCun.
\newblock Spectral networks and locally connected networks on graphs.
\newblock In {\em ICLR}, 2014.

\bibitem{DBLP:journals/tvcg/CaoWZTZ14}
Chen Cao, Yanlin Weng, Shun Zhou, Yiying Tong, and Kun Zhou.
\newblock Facewarehouse: {A} {3D} facial expression database for visual
  computing.
\newblock {\em TVCG}, 20(3):413--425, 2014.

\bibitem{DBLP:conf/nips/DefferrardBV16}
Micha{\"{e}}l Defferrard, Xavier Bresson, and Pierre Vandergheynst.
\newblock Convolutional neural networks on graphs with fast localized spectral
  filtering.
\newblock In {\em NeurIPS}, pages 3837--3845, 2016.

\bibitem{DBLP:journals/pami/DhillonGK07}
Inderjit~S. Dhillon, Yuqiang Guan, and Brian Kulis.
\newblock Weighted graph cuts without eigenvectors {A} multilevel approach.
\newblock {\em PAMI}, 29(11):1944--1957, 2007.

\bibitem{DBLP:conf/cvpr/DoostiNMC20}
Bardia Doosti, Shujon Naha, Majid Mirbagheri, and David~J. Crandall.
\newblock Hope-net: {A} graph-based model for hand-object pose estimation.
\newblock In {\em CVPR}, pages 6607--6616, 2020.

\bibitem{DBLP:journals/tog/EggerSTWZBBBKRT20}
Bernhard Egger, William A.~P. Smith, Ayush Tewari, Stefanie Wuhrer, Michael
  Zollh{\"{o}}fer, Thabo Beeler, Florian Bernard, Timo Bolkart, Adam
  Kortylewski, Sami Romdhani, Christian Theobalt, Volker Blanz, and Thomas
  Vetter.
\newblock {3D} morphable face models - past, present, and future.
\newblock {\em {ACM} TOG}, 39(5):157:1--157:38, 2020.

\bibitem{DBLP:conf/aaai/GaoZZYYY21}
Zhongpai Gao, Guangtao Zhai, Juyong Zhang, Junchi Yan, Yiyan Yang, and Xiaokang
  Yang.
\newblock Learning local neighboring structure for robust {3D} shape
  representation.
\newblock In {\em AAAI}, 2021.

\bibitem{DBLP:conf/siggraph/GarlandH97}
Michael Garland and Paul~S. Heckbert.
\newblock Surface simplification using quadric error metrics.
\newblock In {\em SIGGRAPH}, pages 209--216, 1997.

\bibitem{DBLP:conf/cvpr/GeRLXWCY19}
Liuhao Ge, Zhou Ren, Yuncheng Li, Zehao Xue, Yingying Wang, Jianfei Cai, and
  Junsong Yuan.
\newblock {3D} hand shape and pose estimation from a single {RGB} image.
\newblock In {\em CVPR}, pages 10833--10842, 2019.

\bibitem{DBLP:conf/iccvw/GongCBZ19}
Shunwang Gong, Lei Chen, Michael~M. Bronstein, and Stefanos Zafeiriou.
\newblock Spiralnet++: {A} fast and highly efficient mesh convolution operator.
\newblock In {\em {ICCV} Workshops}, pages 4141--4148, 2019.

\bibitem{DBLP:conf/nips/HamiltonYL17}
William~L. Hamilton, Zhitao Ying, and Jure Leskovec.
\newblock Inductive representation learning on large graphs.
\newblock In {\em NeurIPS}, pages 1024--1034, 2017.

\bibitem{DBLP:journals/tog/HanockaHFGFC19}
Rana Hanocka, Amir Hertz, Noa Fish, Raja Giryes, Shachar Fleishman, and Daniel
  Cohen{-}Or.
\newblock {MeshCNN}: a network with an edge.
\newblock {\em TOG}, 38(4):90:1--90:12, 2019.

\bibitem{DBLP:journals/corr/HenaffBL15}
Mikael Henaff, Joan Bruna, and Yann LeCun.
\newblock Deep convolutional networks on graph-structured data.
\newblock {\em CoRR}, abs/1506.05163, 2015.

\bibitem{DBLP:conf/iclr/KipfW17}
Thomas~N. Kipf and Max Welling.
\newblock Semi-supervised classification with graph convolutional networks.
\newblock In {\em ICLR}, 2017.

\bibitem{DBLP:journals/tog/LiWP10}
Hao Li, Thibaut Weise, and Mark Pauly.
\newblock Example-based facial rigging.
\newblock {\em TOG}, 29(4):32:1--32:6, 2010.

\bibitem{DBLP:conf/aaai/LiWZH18}
Ruoyu Li, Sheng Wang, Feiyun Zhu, and Junzhou Huang.
\newblock Adaptive graph convolutional neural networks.
\newblock In {\em AAAI}, pages 3546--3553, 2018.

\bibitem{DBLP:journals/tog/LiBBL017}
Tianye Li, Timo Bolkart, Michael~J. Black, Hao Li, and Javier Romero.
\newblock Learning a model of facial shape and expression from {4D} scans.
\newblock {\em TOG}, 36(6):194:1--194:17, 2017.

\bibitem{DBLP:conf/eccv/LimDCK18}
Isaak Lim, Alexander Dielen, Marcel Campen, and Leif Kobbelt.
\newblock A simple approach to intrinsic correspondence learning on
  unstructured {3D} meshes.
\newblock In {\em ECCV}, pages 349--362, 2018.

\bibitem{DBLP:conf/cvpr/LitanyBBM18}
Or Litany, Alexander~M. Bronstein, Michael~M. Bronstein, and Ameesh Makadia.
\newblock Deformable shape completion with graph convolutional autoencoders.
\newblock In {\em CVPR}, pages 1886--1895, 2018.

\bibitem{DBLP:journals/tog/LiuKCAJ20}
Hsueh{-}Ti~Derek Liu, Vladimir~G. Kim, Siddhartha Chaudhuri, Noam Aigerman, and
  Alec Jacobson.
\newblock Neural subdivision.
\newblock {\em TOG}, 39(4):124, 2020.

\bibitem{DBLP:journals/tog/LoperM0PB15}
Matthew Loper, Naureen Mahmood, Javier Romero, Gerard Pons{-}Moll, and
  Michael~J. Black.
\newblock {SMPL:} a skinned multi-person linear model.
\newblock {\em TOG}, 34(6):248:1--248:16, 2015.

\bibitem{DBLP:conf/3dim/MalikENVTHS18}
Jameel Malik, Ahmed Elhayek, Fabrizio Nunnari, Kiran Varanasi, Kiarash
  Tamaddon, Alexis H{\'{e}}loir, and Didier Stricker.
\newblock Deephps: End-to-end estimation of {3D} hand pose and shape by
  learning from synthetic depth.
\newblock In {\em International Conference on 3D Vision, 3DV}, pages 110--119,
  2018.

\bibitem{DBLP:conf/iccvw/MasciBBV15}
Jonathan Masci, Davide Boscaini, Michael~M. Bronstein, and Pierre
  Vandergheynst.
\newblock Geodesic convolutional neural networks on riemannian manifolds.
\newblock In {\em {ICCV} Workshops}, pages 832--840, 2015.

\bibitem{DBLP:conf/cvpr/MontiBMRSB17}
Federico Monti, Davide Boscaini, Jonathan Masci, Emanuele Rodol{\`{a}}, Jan
  Svoboda, and Michael~M. Bronstein.
\newblock Geometric deep learning on graphs and manifolds using mixture model
  {CNNs}.
\newblock In {\em CVPR}, pages 5425--5434, 2017.

\bibitem{DBLP:conf/eccv/OsmanBB20}
Ahmed A.~A. Osman, Timo Bolkart, and Michael~J. Black.
\newblock {STAR:} sparse trained articulated human body regressor.
\newblock In {\em ECCV}, pages 598--613, 2020.

\bibitem{DBLP:conf/avss/PaysanKARV09}
Pascal Paysan, Reinhard Knothe, Brian Amberg, Sami Romdhani, and Thomas Vetter.
\newblock A {3D} face model for pose and illumination invariant face
  recognition.
\newblock In {\em {IEEE} International Conference on Advanced Video and Signal
  Based Surveillance, {AVSS}}, pages 296--301, 2009.

\bibitem{DBLP:conf/cvpr/Ploumpis0PSZ19}
Stylianos Ploumpis, Haoyang Wang, Nick~E. Pears, William A.~P. Smith, and
  Stefanos Zafeiriou.
\newblock Combining {3D} morphable models: {A} large scale face-and-head model.
\newblock In {\em CVPR}, pages 10934--10943, 2019.

\bibitem{DBLP:conf/eccv/RanjanBSB18}
Anurag Ranjan, Timo Bolkart, Soubhik Sanyal, and Michael~J. Black.
\newblock Generating {3D} faces using convolutional mesh autoencoders.
\newblock In {\em ECCV}, pages 725--741, 2018.

\bibitem{DBLP:journals/tog/0002TB17}
Javier Romero, Dimitrios Tzionas, and Michael~J. Black.
\newblock Embodied hands: modeling and capturing hands and bodies together.
\newblock {\em TOG}, 36(6):245:1--245:17, 2017.

\bibitem{DBLP:conf/cvpr/SchultEKL20}
Jonas Schult, Francis Engelmann, Theodora Kontogianni, and Bastian Leibe.
\newblock Dualconvmesh-net: Joint geodesic and euclidean convolutions on {3D}
  meshes.
\newblock In {\em CVPR}, pages 8609--8619, 2020.

\bibitem{DBLP:conf/ipmi/StynerRNZSTD03}
Martin Styner, Kumar~T. Rajamani, Lutz{-}Peter Nolte, Gabriel Zsemlye,
  G{\'{a}}bor Sz{\'{e}}kely, Christopher~J. Taylor, and Rhodri~H. Davies.
\newblock Evaluation of {3D} correspondence methods for model building.
\newblock In {\em International Conference on Information Processing in Medical
  Imaging {IPMI}}, pages 63--75, 2003.

\bibitem{DBLP:conf/nips/SukhbaatarSWF15}
Sainbayar Sukhbaatar, Arthur Szlam, Jason Weston, and Rob Fergus.
\newblock End-to-end memory networks.
\newblock In {\em NeurIPS}, pages 2440--2448, 2015.

\bibitem{DBLP:conf/cvpr/TaigmanYRW14}
Yaniv Taigman, Ming Yang, Marc'Aurelio Ranzato, and Lior Wolf.
\newblock Deepface: Closing the gap to human-level performance in face
  verification.
\newblock In {\em CVPR}, pages 1701--1708, 2014.

\bibitem{DBLP:conf/cvpr/Tan0LX18}
Qingyang Tan, Lin Gao, Yu{-}Kun Lai, and Shihong Xia.
\newblock Variational autoencoders for deforming {3D} mesh models.
\newblock In {\em CVPR}, pages 5841--5850, 2018.

\bibitem{DBLP:conf/iccv/TewariZK0BPT17}
Ayush Tewari, Michael Zollh{\"{o}}fer, Hyeongwoo Kim, Pablo Garrido, Florian
  Bernard, Patrick P{\'{e}}rez, and Christian Theobalt.
\newblock Mofa: Model-based deep convolutional face autoencoder for
  unsupervised monocular reconstruction.
\newblock In {\em {ICCV}}, pages 3735--3744, 2017.

\bibitem{DBLP:journals/tog/ThiesZNVST15}
Justus Thies, Michael Zollh{\"{o}}fer, Matthias Nie{\ss}ner, Levi Valgaerts,
  Marc Stamminger, and Christian Theobalt.
\newblock Real-time expression transfer for facial reenactment.
\newblock {\em TOG}, 34(6):183:1--183:14, 2015.

\bibitem{DBLP:conf/cvpr/TranHMPNM18}
Anh~Tuan Tran, Tal Hassner, Iacopo Masi, Eran Paz, Yuval Nirkin, and
  G{\'{e}}rard~G. Medioni.
\newblock Extreme {3D} face reconstruction: Seeing through occlusions.
\newblock In {\em CVPR}, pages 3935--3944, 2018.

\bibitem{DBLP:conf/eccv/TretschkTZGT20}
Edgar Tretschk, Ayush Tewari, Michael Zollh{\"{o}}fer, Vladislav Golyanik, and
  Christian Theobalt.
\newblock {DEMEA:} deep mesh autoencoders for non-rigidly deforming objects.
\newblock In {\em ECCV}, pages 601--617, 2020.

\bibitem{DBLP:conf/nips/VaswaniSPUJGKP17}
Ashish Vaswani, Noam Shazeer, Niki Parmar, Jakob Uszkoreit, Llion Jones,
  Aidan~N. Gomez, Lukasz Kaiser, and Illia Polosukhin.
\newblock Attention is all you need.
\newblock In {\em NeurIPS}, pages 5998--6008, 2017.

\bibitem{DBLP:conf/iclr/VelickovicCCRLB18}
Petar Velickovic, Guillem Cucurull, Arantxa Casanova, Adriana Romero, Pietro
  Li{\`{o}}, and Yoshua Bengio.
\newblock Graph attention networks.
\newblock In {\em ICLR}, 2018.

\bibitem{DBLP:journals/corr/VermaBV17}
Nitika Verma, Edmond Boyer, and Jakob Verbeek.
\newblock Dynamic filters in graph convolutional networks.
\newblock {\em CoRR}, abs/1706.05206, 2017.

\bibitem{DBLP:journals/corr/abs-1904-05562}
Huawei Wei, Shuang Liang, and Yichen Wei.
\newblock {3D} dense face alignment via graph convolution networks.
\newblock {\em CoRR}, abs/1904.05562, 2019.

\bibitem{DBLP:journals/tog/WiersmaEH20}
Ruben Wiersma, Elmar Eisemann, and Klaus Hildebrandt.
\newblock {CNNs} on surfaces using rotation-equivariant features.
\newblock {\em TOG}, 39(4):92, 2020.

\bibitem{DBLP:journals/tog/YangWSBM11}
Fei Yang, Jue Wang, Eli Shechtman, Lubomir~D. Bourdev, and Dimitris~N. Metaxas.
\newblock Expression flow for {3D}-aware face component transfer.
\newblock {\em TOG}, 30(4):60, 2011.

\bibitem{DBLP:conf/nips/YingY0RHL18}
Zhitao Ying, Jiaxuan You, Christopher Morris, Xiang Ren, William~L. Hamilton,
  and Jure Leskovec.
\newblock Hierarchical graph representation learning with differentiable
  pooling.
\newblock In {\em NeurIPS}, pages 4805--4815, 2018.

\bibitem{DBLP:conf/cvpr/YuanL0DF020}
Yu{-}Jie Yuan, Yu{-}Kun Lai, Jie Yang, Qi Duan, Hongbo Fu, and Lin Gao.
\newblock Mesh variational autoencoders with edge contraction pooling.
\newblock In {\em CVPR Workshops}, pages 1105--1112, 2020.

\bibitem{DBLP:conf/nips/ZhouWLCYSLS20}
Yi Zhou, Chenglei Wu, Zimo Li, Chen Cao, Yuting Ye, Jason~M. Saragih, Hao Li,
  and Yaser Sheikh.
\newblock Fully convolutional mesh autoencoder using efficient spatially
  varying kernels.
\newblock In {\em NeurIPS}, 2020.

\end{thebibliography}
}

\newpage
\appendix

In this supplementary material, we present more details about the experiments in the main paper and show additional experimental results to evaluate our proposed feature aggregation method.
In Sec.~\ref{sec_arch_detail}, we provide details of our evaluated architecture.
Sec.~\ref{sec_dataset_detail} provides the statistical information of the datasets.
Sec.~\ref{sec_implementation_detail} contains implementation details to train the models.
Sec.~\ref{sec_more_exp_results} presents more results about the experiments in the main paper.
In Sec.~\ref{sec_sup_exp_arch}, we give the experimental results on different network architectures.
In Sec.~\ref{sec_sup_exp_ablation}, we show further ablation study experiments on the attention mechanism.
In Sec.~\ref{sec_sup_exp_parameter_sensitivity}, we provide experimental results of parameter sensitivity studies.
We give the model complexity analysis in Sec.~\ref{sec_sup_discussion}.

\section{Architecture details}\label{sec_arch_detail}
Our model consists of an encoder and a decoder. The architecture details of the encoder and decoder are listed in Tables~\ref{table_enc} and~\ref{table_dec}, resepectively.
The convolution is Chebyshev convolution filter with $K=6$ Chebyshev polynomials for CoMA and spiral convolution of 1 hop for Neural3DMM.
The aggregation, including downsampling and upsampling, is either implemented by QEM as in~\cite{DBLP:conf/eccv/RanjanBSB18} or accomplished by our proposed attention based module.
The dimension of latent representation $n_z$ is set as one of $\{8,16,32,64\}$ in the evaluated settings.
The numbers of vertices at hierarchical levels are summarized in Table~\ref{table_num_vertex}.

\section{Datset statistics}\label{sec_dataset_detail}
Table~\ref{table_dataset} summarizes the statistics of the datasets used for evaluating 3D models. Since the deformations are randomly generated, there is no identity and pose category information for the SYNHAND dataset.

\setlength{\tabcolsep}{4pt}
\begin{table}
\begin{center}
\caption{Architecture of the encoder}
\label{table_enc}
\begin{tabular}{lcc}
layer           & input size    & output size   \\ \hline
convolution     & $n_4$*3        & $n_4$*16     \\
downsampling    & $n_4$*16       & $n_3$*16     \\
convolution     & $n_3$*16        & $n_3$*16     \\
downsampling    & $n_3$*16       & $n_2$*16     \\
convolution     & $n_2$*16        & $n_2$*16     \\
downsampling    & $n_2$*16       & $n_1$*16     \\
convolution     & $n_1$*16        & $n_1$*32     \\
downsampling    & $n_1$*32       & $n_0$*32     \\
fully connected & $n_0$*32      & $n_z$
\end{tabular}
\end{center}
\end{table}
\setlength{\tabcolsep}{4pt}
\begin{table}
\begin{center}
\caption{Architecture of the decoder}
\label{table_dec}
\begin{tabular}{lcc}
layer           & input size    & output size   \\ \hline
fully connected & $n_z$         & $n_0$*32      \\
upsampling    & $n_0$*32       & $n_1$*32     \\
convolution     & $n_1$*32        & $n_1$*32     \\
upsampling    & $n_1$*32       & $n_2$*32     \\
convolution     & $n_2$*32        & $n_2$*16     \\
upsampling    & $n_2$*16       & $n_3$*16     \\
convolution     & $n_3$*16        & $n_3$*16     \\
upsampling    & $n_3$*16       & $n_4$*16     \\
convolution     & $n_4$*16        & $n_4$*16    \\
convolution     & $n_4$*16        & $n_4$*3     
\end{tabular}
\end{center}
\end{table}
\setlength{\tabcolsep}{4pt}
\begin{table}
\begin{center}
\caption{Number of vertices at hierarchical levels}
\label{table_num_vertex}
\begin{tabular}{lccccc}
dataset     & $n_0$     & $n_1$     & $n_2$     & $n_3$     & $n_4$     \\ \hline
COMA~\cite{DBLP:conf/eccv/RanjanBSB18}        & 20        & 79        & 314       & 1256      & 5023      \\
DFAUST~\cite{DBLP:conf/cvpr/Bogo0PB17}      & 27        & 108       & 431       & 1723      & 6890      \\
SYNHAND~\cite{DBLP:conf/3dim/MalikENVTHS18}     & 5         & 19        & 75        & 299       & 1193
\end{tabular}
\end{center}
\end{table}
\setlength{\tabcolsep}{4pt}
\begin{table}
\begin{center}
\caption{Summary of statistics of the benchmark datasets }
\label{table_dataset}
\begin{tabular}{lcccc}
name        & \#mesh     & \#vertex     & \#ID    & \#pose/expression     \\ \hline
COMA        & 20K        & 5023        & 12       & 12      \\
DFAUST      & 40K        & 6890       & 10       & 14      \\
SYNHAND     & 100K         & 1193        & -        & -
\end{tabular}
\end{center}
\end{table}

\begin{figure*}[tb]
\centering
{
    \begin{minipage}[t]{5.4cm}
    \centering
    \includegraphics[width=5.4cm]{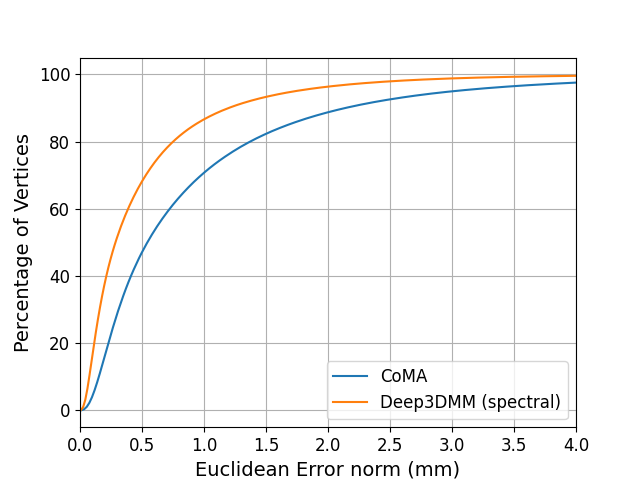}
    \end{minipage}
}
\hspace{0.1cm}
{
    \begin{minipage}[t]{5.4cm}
    \includegraphics[width=5.4cm]{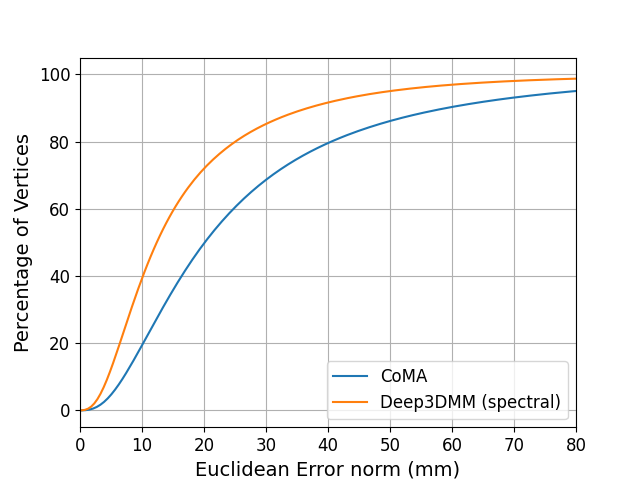}
    \end{minipage}
}
\hspace{0.1cm}
{
    \begin{minipage}[t]{5.4cm}
    \includegraphics[width=5.4cm]{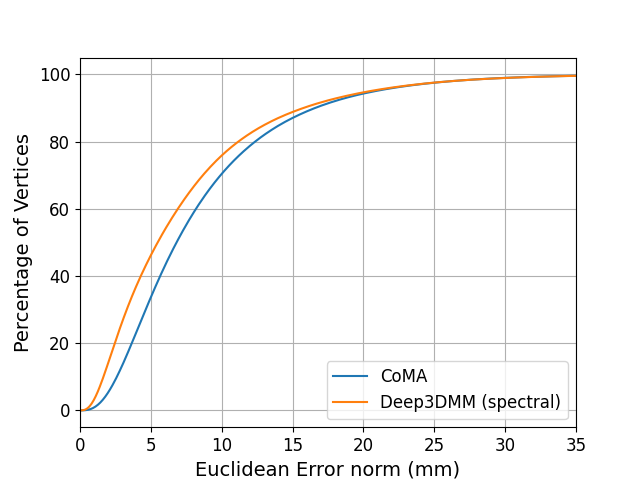}
    \end{minipage}
}
\caption{Cumulative Euclidean errors of CoMA method and our Deep3DMM (spectral) method with proposed feature aggregation module on COMA(left), DFAUST(middle), and SYNHAND(right) datasets}
\label{fig_cdf}
\end{figure*}

\begin{figure*}[h]
\centering
\includegraphics[width=.8\linewidth]{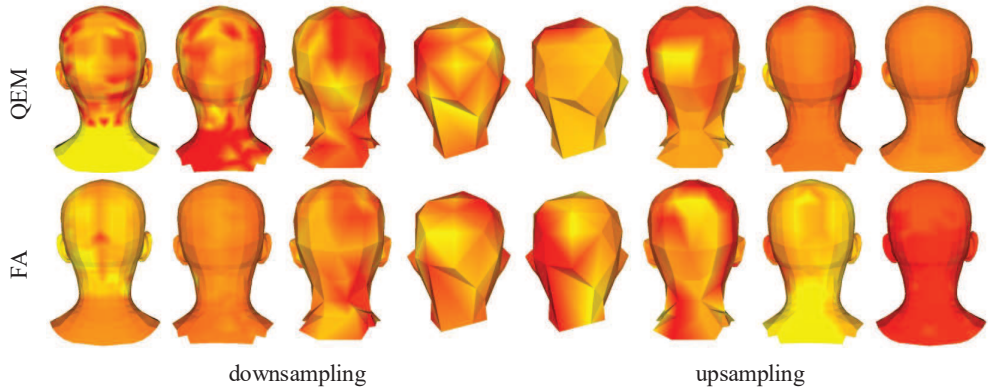}
\caption{Back view of the visualization of mapping matrices of down-sampling and up-sampling on COMA dataset with t-SNE (best viewed in color). The first and second rows are the results of quadric error minimization in~\cite{DBLP:conf/eccv/RanjanBSB18,DBLP:conf/iccv/BouritsasBPZB19} and our proposed feature aggregation, respectively }
\label{fig_viz_mapping_hierarchy_coma_back}
\end{figure*}

\section{Implementation details}\label{sec_implementation_detail}
We implement the models with PyTorch.
We adopt the training settings suggested by the original authors~\cite{DBLP:conf/eccv/RanjanBSB18,DBLP:conf/iccv/BouritsasBPZB19}.
We train the CoMA and Deep3DMM (spectral)  models for 300 epochs with learning rate of 8e-3.
We train the Neural3DMM and Deep3DMM (spiral)  models for 200 epochs with learning rate of 1e-3.

\begin{figure*}[h]
\centering
{
    \begin{minipage}[t]{3.8cm}
    \centering
    \includegraphics[width=3.8cm]{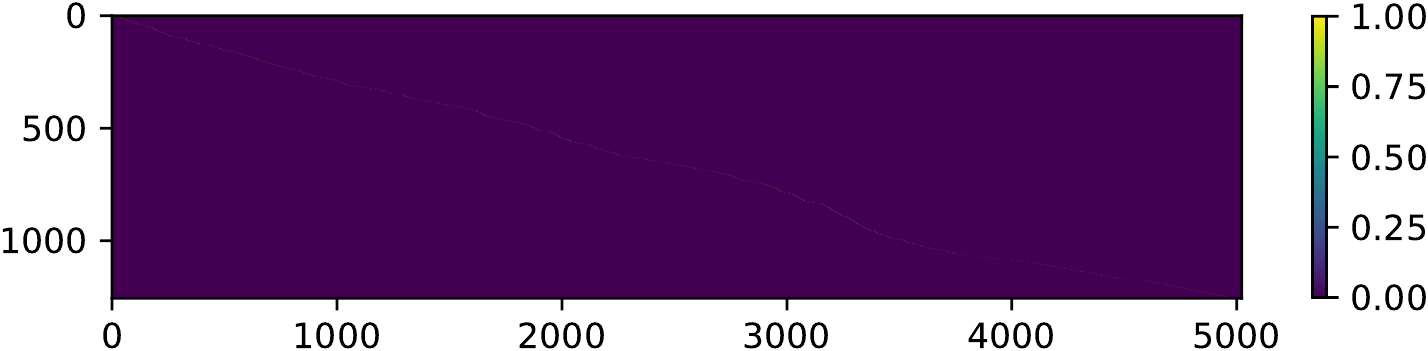}
    \end{minipage}
}
\hspace{0.1cm}
\centering
{
    \begin{minipage}[t]{3.8cm}
    \centering
    \includegraphics[width=3.8cm]{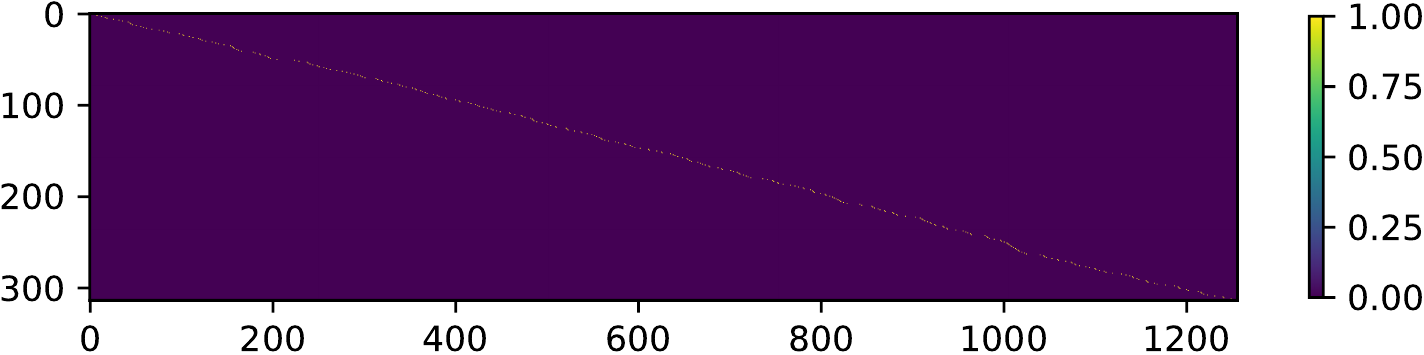}
    \end{minipage}
}
\hspace{0.1cm}
\centering
{
    \begin{minipage}[t]{3.8cm}
    \centering
    \includegraphics[width=3.8cm]{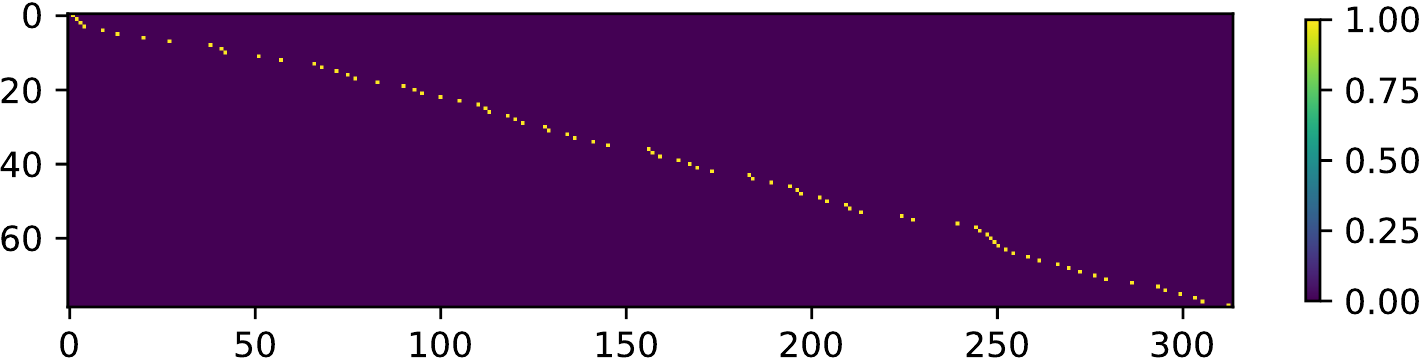}
    \end{minipage}
}
\hspace{0.1cm}
\centering
{
    \begin{minipage}[t]{3.8cm}
    \centering
    \includegraphics[width=3.8cm]{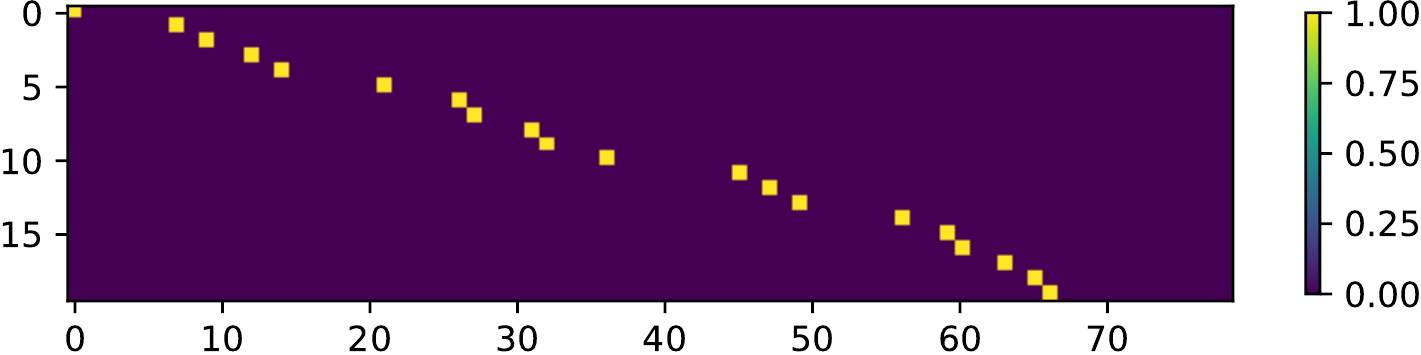}
    \end{minipage}
}

{
    \begin{minipage}[t]{3.8cm}
    \centering
    \includegraphics[width=3.8cm]{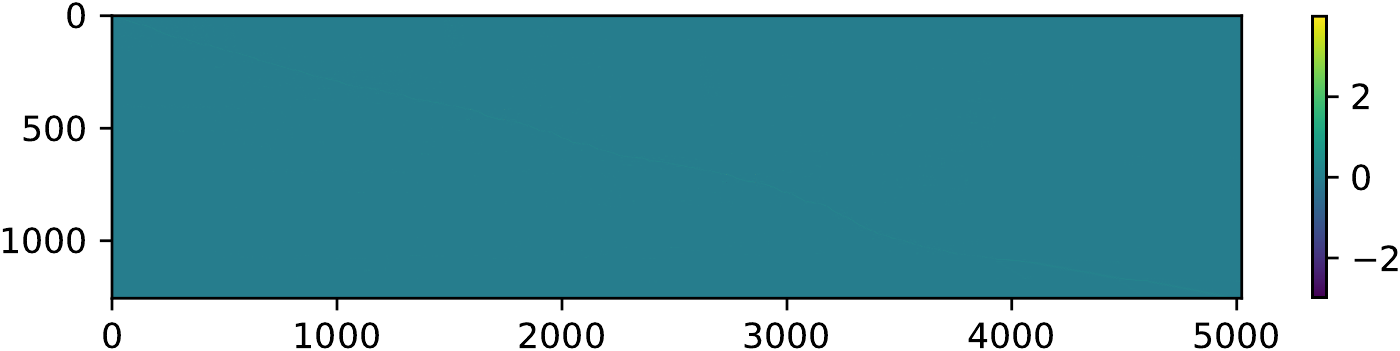}
    \end{minipage}
}
\hspace{0.1cm}
\centering
{
    \begin{minipage}[t]{3.8cm}
    \centering
    \includegraphics[width=3.8cm]{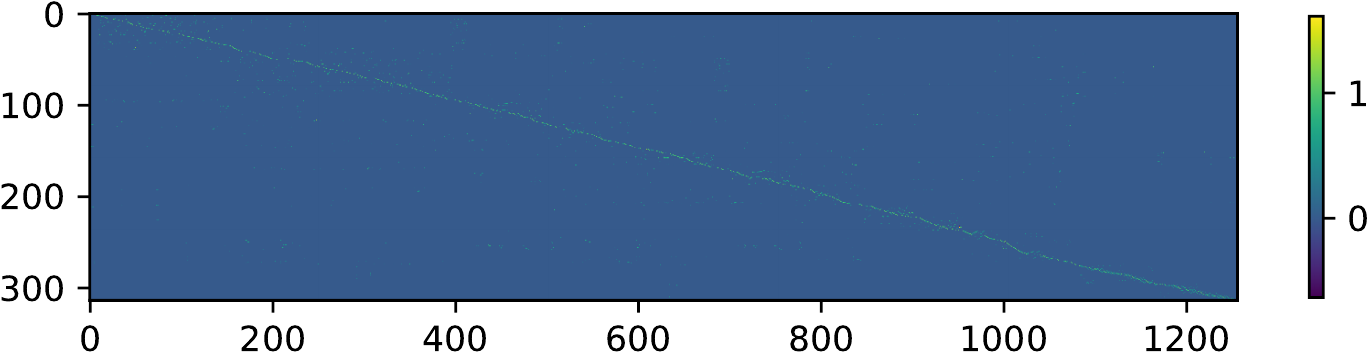}
    \end{minipage}
}
\hspace{0.1cm}
\centering
{
    \begin{minipage}[t]{3.8cm}
    \centering
    \includegraphics[width=3.8cm]{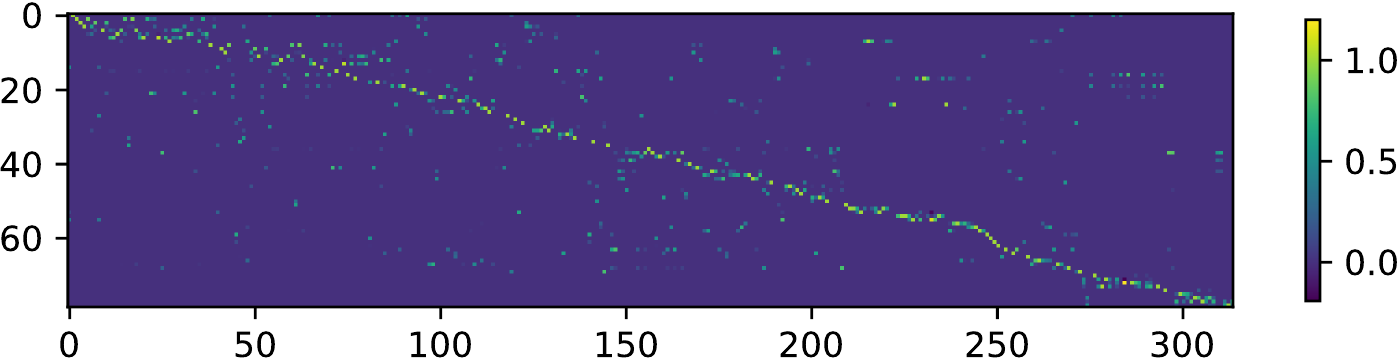}
    \end{minipage}
}
\hspace{0.1cm}
\centering
{
    \begin{minipage}[t]{3.8cm}
    \centering
    \includegraphics[width=3.8cm]{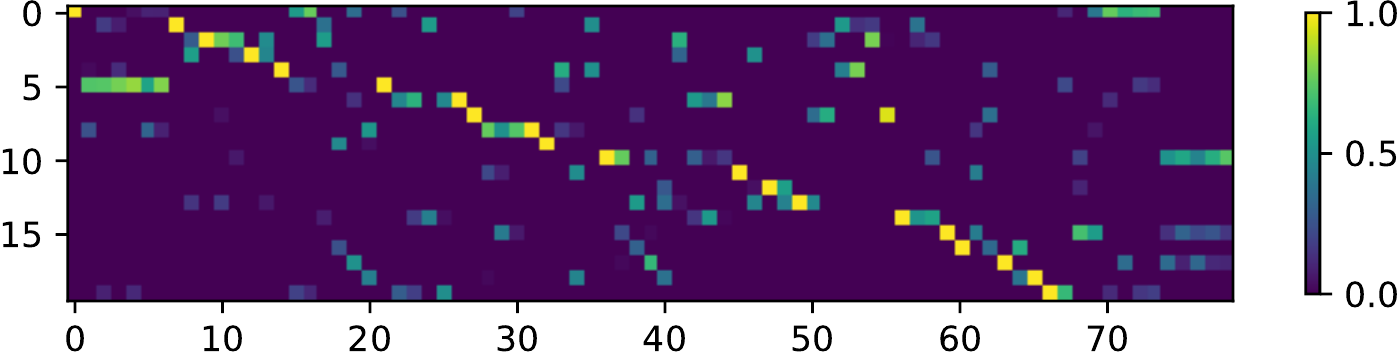}
    \end{minipage}
}
\caption{Visualization of mapping matrices of downsampling (top row) and upsampling (bottom row) with existing QEM method on COMA dataset(best viewed in color and zoom in to see details)}
\label{fig_heatmap_coma}
\end{figure*}

\begin{figure*}[!h]
\centering
{
    \begin{minipage}[t]{3.8cm}
    \centering
    \includegraphics[width=3.8cm]{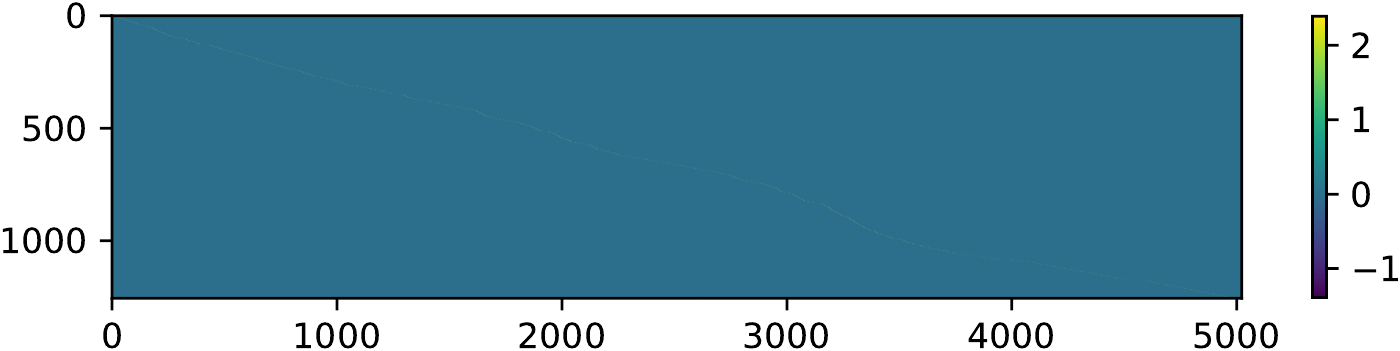}
    \end{minipage}
}
\hspace{0.1cm}
\centering
{
    \begin{minipage}[t]{3.8cm}
    \centering
    \includegraphics[width=3.8cm]{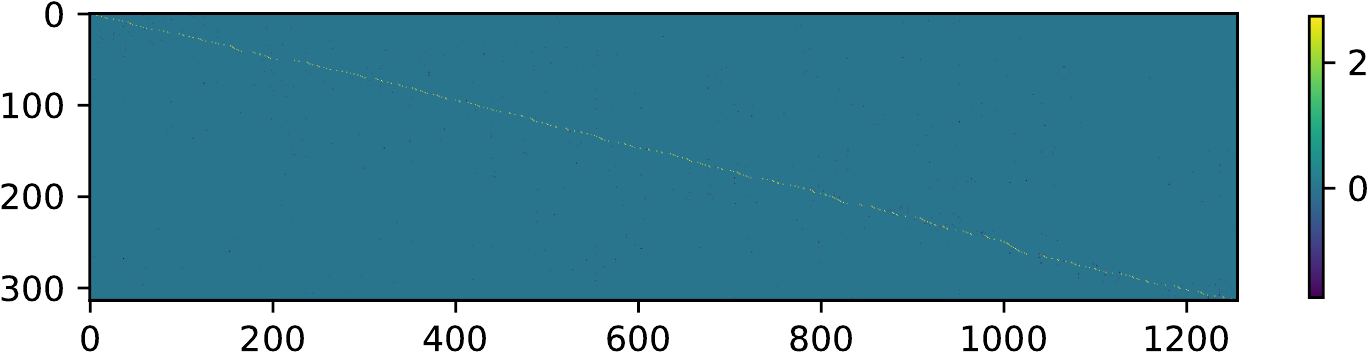}
    \end{minipage}
}
\hspace{0.1cm}
\centering
{
    \begin{minipage}[t]{3.8cm}
    \centering
    \includegraphics[width=3.8cm]{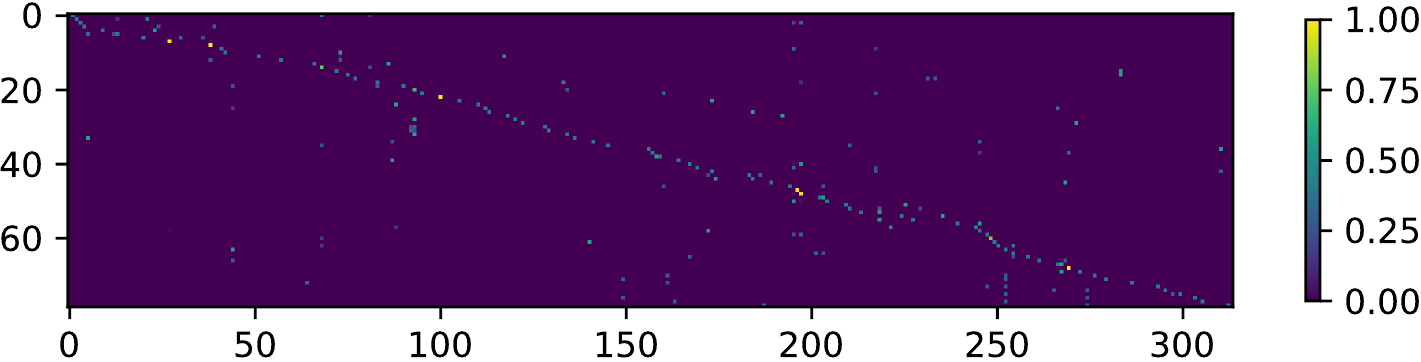}
    \end{minipage}
}
\hspace{0.1cm}
\centering
{
    \begin{minipage}[t]{3.8cm}
    \centering
    \includegraphics[width=3.8cm]{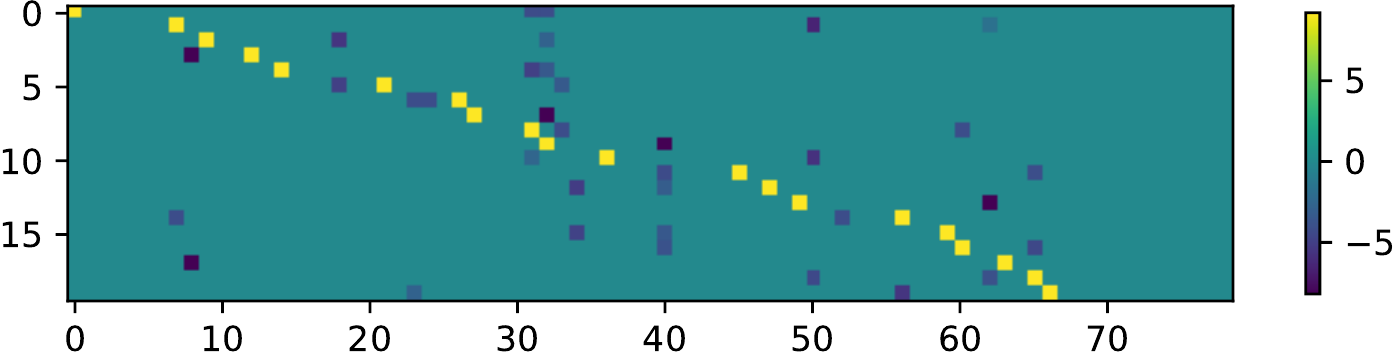}
    \end{minipage}
}

{
    \begin{minipage}[t]{3.8cm}
    \centering
    \includegraphics[width=3.8cm]{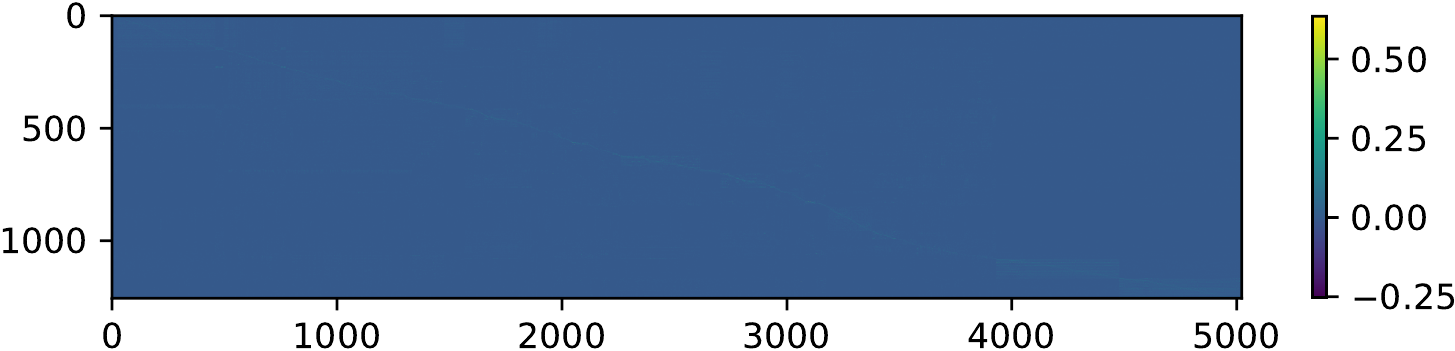}
    \end{minipage}
}
\hspace{0.1cm}
\centering
{
    \begin{minipage}[t]{3.8cm}
    \centering
    \includegraphics[width=3.8cm]{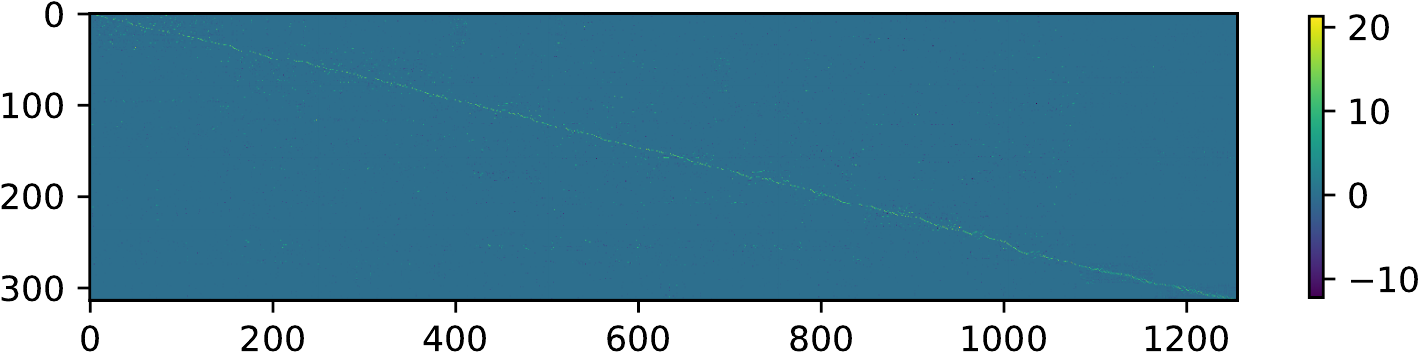}
    \end{minipage}
}
\hspace{0.1cm}
\centering
{
    \begin{minipage}[t]{3.8cm}
    \centering
    \includegraphics[width=3.8cm]{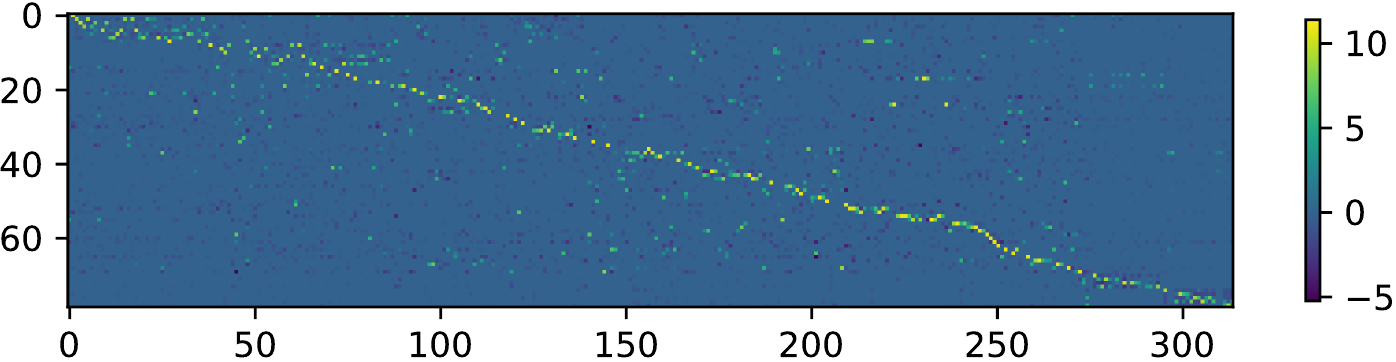}
    \end{minipage}
}
\hspace{0.1cm}
\centering
{
    \begin{minipage}[t]{3.8cm}
    \centering
    \includegraphics[width=3.8cm]{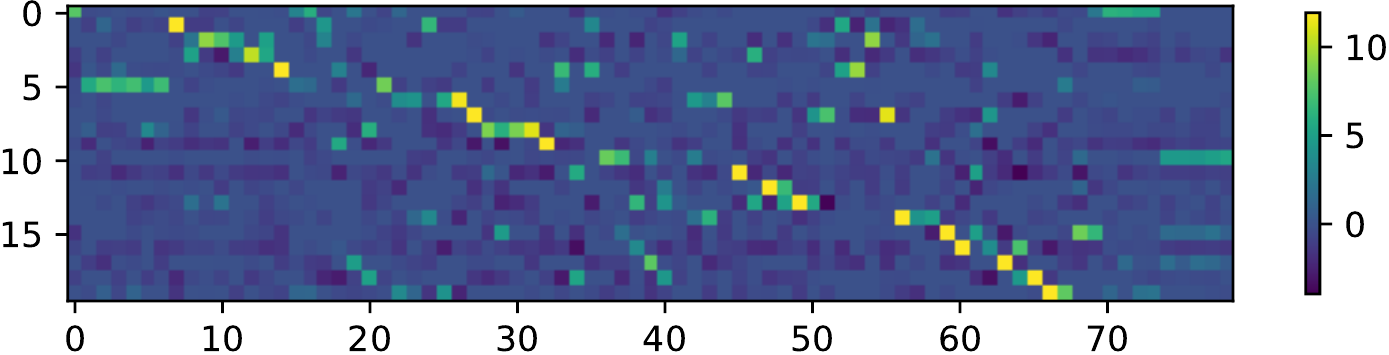}
    \end{minipage}
}
\caption{Visualization of mapping matrices of downsampling (top row) and upsampling (bottom row) with our proposed feature aggregation module on COMA dataset(best viewed in color and zoom in to see details)}
\label{fig_heatmap_coma_fa}
\end{figure*}

\begin{figure}[tb]
\centering
\includegraphics[width=.98\linewidth]{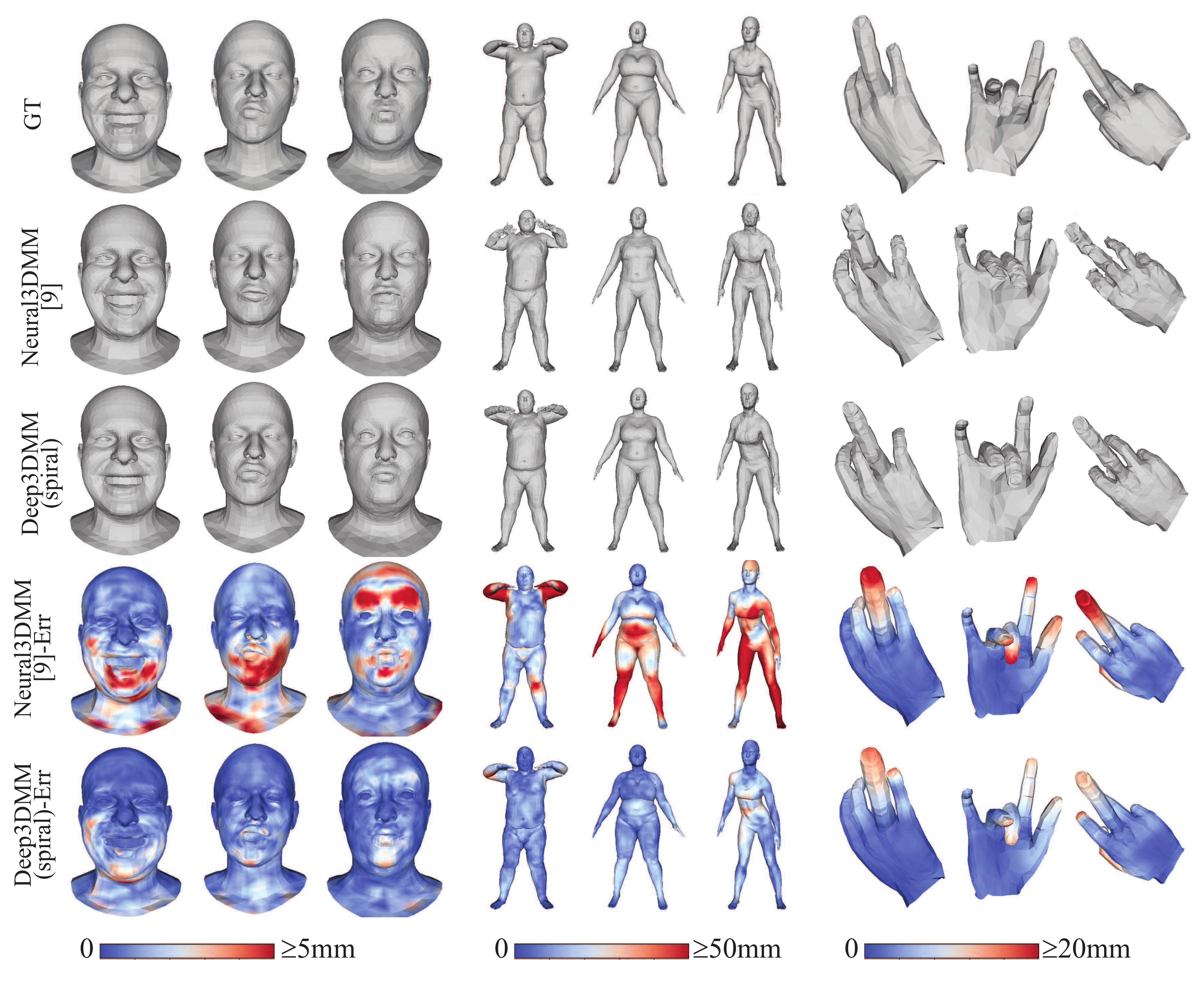}
\caption{Qualitative comparison of spiral convolution based models on COMA (left), DFAUST (middle), and SYNHAND (right) datasets. The first row is the ground truth shapes. The second and third rows show the reconstructed shapes, while the fourth and fifth rows show the corresponding reconstruction errors}
\label{fig_viz_recon_spiral}
\end{figure}

\section{More experimental results}\label{sec_more_exp_results}
In this section, we provide more experimental results of the reconstruction errors and show more visualizations of the mapping matrices.

\subsection{Cumulative Euclidean errors}

\par
In Fig.~\ref{fig_cdf}, we show the cumulative distribution of the Euclidean errors with and without our feature aggregation module for CoMA model with latent dimension of 8.
We can find that for a given error threshold, more vertices can satisfy the constraint with lower error by applying our feature aggregation module. This is consistent with the observation from the qualitative results in the main paper.

\subsection{More Visualization of the mapping matrices}
In Fig.~\ref{fig_viz_mapping_hierarchy_coma_back}, we show the back view of the mapping matrices on COMA dataset, which is complementary to the front view in the main paper. The pattern is similar to the front view in the main paper.
In Figs.~\ref{fig_heatmap_coma} and~\ref{fig_heatmap_coma_fa}, we directly show the values of the mapping matrices on COMA dataset.
Since there are large number of columns and rows in each mapping matrix, a better view can be obtained by zooming in on the figures.
While the mapping matrices obtained by QEM and our feature aggregation mechanism demonstrate similar pattern in positions of the dominant elements, the values of these elements are different for these two methods.
This is because our proposed  feature aggregation mechanism enables learning the weights from the training data automatically.
Moreover, the mapping matrices learned by FA are more dense than those computed by QEM. This shows that our proposed  feature aggregation mechanism also learns the receptive fields.

\subsection{Qualitative results with spiral convolutions}
In Fig.~\ref{fig_viz_recon_spiral}, we show the per vertex Euclidean error of different morphable models on several shapes from the three datasets for qualitative comparison. The latent dimension is set as 8.
We can find that our model can reduce the large errors of the compare model (red regions) by  providing accurate predictions.
Our model can also recover more details than the compared model, leading to more realistic shapes and lower reconstruction errors over almost all regions.

\setlength{\tabcolsep}{4pt}
\begin{table*}
\begin{center}
\caption{Reconstruction errors on extrapolation setting}
\label{table_extrapolation}
\begin{tabular}{l|cc|cc|cc|cc}
 & \multicolumn{2}{c|}{Deep3DMM(spectral)}	& \multicolumn{2}{c|}{CoMA~\cite{DBLP:conf/eccv/RanjanBSB18}}	& \multicolumn{2}{c|}{PCA} & \multicolumn{2}{c}{FLAME~\cite{DBLP:journals/tog/LiBBL017}} \\
Sequence & Mean Error & Median  & Mean Error & Median  & Mean Error & Median  & Mean Error & Median \\\hline
bareteeth  & \textbf{1.190$\pm$1.524} & \textbf{0.691}   & 1.376$\pm$1.536 & 0.856    & 1.957$\pm$1.888 & 1.335       & 2.002$\pm$1.456 & 1.606 \\
cheeks in  & \textbf{1.071$\pm$1.322} & \textbf{0.646}   & 1.288$\pm$1.501 & 0.794    & 1.854$\pm$1.906 & 1.179       & 2.011$\pm$1.468 & 1.609 \\
eyebrow  & \textbf{0.851$\pm$1.011} & \textbf{0.505}   & 1.053$\pm$1.088 & 0.706    & 1.609$\pm$1.535 & 1.090       & 1.862$\pm$1.342 & 1.516 \\
high smile  & \textbf{1.037$\pm$1.164} & \textbf{0.614}   & 1.205$\pm$1.252 & 0.772    & 1.841$\pm$1.831 & 1.246       & 1.960$\pm$1.370 & 1.625 \\
lips back  & \textbf{1.060}$\pm$1.590 & \textbf{0.580}   & 1.193$\pm$\textbf{1.476} & 0.708    & 1.842$\pm$1.947 & 1.198       & 2.047$\pm$1.485 & 1.639 \\
lips up  & \textbf{0.902$\pm$1.114} & \textbf{0.497}   & 1.081$\pm$1.192 & 0.656    & 1.788$\pm$1.764 & 1.216       & 1.983$\pm$1.427 & 1.616 \\
mouth down  & \textbf{0.847$\pm$1.062} & \textbf{0.517}   & 1.050$\pm$1.183 & 0.654    & 1.618$\pm$1.594 & 1.105       & 2.029$\pm$1.454 & 1.651 \\
mouth extreme  & \textbf{1.139$\pm$1.468} & \textbf{0.640}   & 1.336$\pm$1.820 & 0.738    & 2.011$\pm$2.405 & 1.224       & 2.028$\pm$1.464 & 1.613 \\
mouth middle  & \textbf{0.745$\pm$0.934} & \textbf{0.439}   & 1.017$\pm$1.192 & 0.610    & 1.697$\pm$1.715 & 1.133       & 2.043$\pm$1.496 & 1.620 \\
mouth open  & \textbf{0.741$\pm$0.996} & \textbf{0.431}   & 0.961$\pm$1.127 & 0.583    & 1.612$\pm$1.728 & 1.060       & 1.894$\pm$1.433 & 1.544 \\
mouth side  & \textbf{1.103}$\pm$1.711 & \textbf{0.567}   & 1.264$\pm$\textbf{1.611} & 0.730    & 1.894$\pm$2.274 & 1.132       & 2.090$\pm$1.510 & 1.659 \\
mouth up  & \textbf{0.835$\pm$0.983} & \textbf{0.501}   & 1.097$\pm$1.212 & 0.683    & 1.710$\pm$1.680 & 1.159       & 2.067$\pm$1.485 & 1.680 \\
\end{tabular}
\end{center}
\end{table*}

\subsection{Extrapolation experiment}
To further measure the generalization capability of our model, we follow the setting in~\cite{DBLP:conf/eccv/RanjanBSB18} to reconstruct expressions that are excluded from the training set. We conduct 12 different experiments to evaluate the performance on each of the 12 expressions when the remaining 11 expressions are used as the training samples. We compare the results (mean, standard deviation and median of the Euclidean distance) of our model with CoMA~\cite{DBLP:conf/eccv/RanjanBSB18}, PCA and FLAME~\cite{DBLP:journals/tog/LiBBL017}, as shown in Table~\ref{table_extrapolation}. The results of the compared methods are taken from~\cite{DBLP:conf/eccv/RanjanBSB18}. We can observe that our model achieve better performance than the compared methods on all expression sequences.

\setlength{\tabcolsep}{4pt}
\begin{table}
\begin{center}
\caption{Reconstruction errors with different settings of convolution filters }
\label{table_filter}
\begin{tabular}{l|cc}
method      & simple    & wider    \\\hline
Neural3DMM      & 0.785 & 0.525         \\ 
Deep3DMM (spiral)   & 0.487     & 0.420
\end{tabular}
\end{center}
\end{table}

\section{Results with different network architectures}\label{sec_sup_exp_arch}
In this section, we evaluate our proposed feature aggregation module on different variants of network architecture, including the number of convolution filters of the spiral convolution and the Chebyshev polynomial order of the spectral convolution.
\subsection{Number of convolution filters}
To explore the effectiveness of our proposed feature aggregation module with different network architectures, we conduct experiments on two settings of the number of convolution filters.
The simple setting denotes the network architecture introduced in Tables~\ref{table_enc} and~\ref{table_dec}, where the number of filters are (3,16,16,16,32) and (32,32,16,16,16,3) for the encoder and decoder, respectively.
The wider setting denotes a larger number of filters, where they are (3,16,32,64,128) and (128,64,32,32,16,3) for the encoder and decoder, respectively.
Table~\ref{table_filter} shows the reconstruction errors on COMA dataset with the latent dimension of 8 for Neural3DMM and out spiral convolution based Deep3DMM.
We can see that our model performs better than the baseline model in both scenarios.
Note that our model has the same inference parameters as the compared model. Our model only introduces $8+(5023+1256+1256+314+314+79+79+20)\times c$ parameters for the keys and querys at the training stage.

\subsection{Chebyshev polynomial order}
In Table~\ref{table_K}, we show the results for variant Chebshev polynomial order K. The experiments are again conducted on COMA dataset with the latent dimension as 8.
We can see that model with our feature aggregation module can consistently perform better.

\setlength{\tabcolsep}{4pt}
\begin{table}
\begin{center}
\caption{Reconstruction errors with different order K of Chebyshev polynomial}
\label{table_K}


\begin{tabular}{l|cc}
method      & CoMA    & Deep3DMM (spectral)    \\\hline
K=6      & 0.939 & 0.519         \\ 
K=3   & 1.031     & 0.558
\end{tabular}
\end{center}
\end{table}

\section{Ablation studies}\label{sec_sup_exp_ablation}
In this section, we provide ablation studies to show the effect of each component in the model.
The experiments are conducted on COMA dataset by using the spectral convolution with the latent dimension as 8.

\subsection{Effect of the topk selection}
\begin{figure}[tb]
\centering
\includegraphics[width=.7\linewidth]{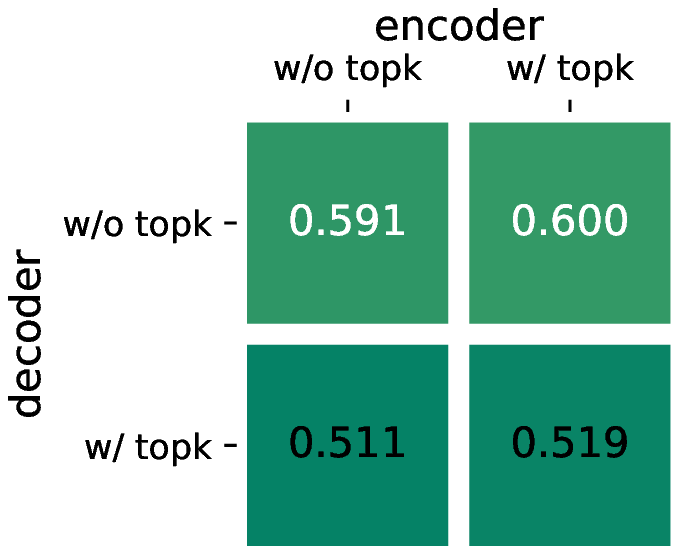}
\caption{Reconstruction errors w/ and w/o applying the topk selection in FA module}
\label{fig_coma_ablation_topk}
\end{figure}

In Fig.~\ref{fig_coma_ablation_topk}, we show the reconstruction errors with and without the topk selection for the mask operation in the feature aggregation module.
As we can see, the error is reduced by applying the topk selection strategy in the decoder by a large margin.
And the performance is slightly deteriorated by applying the topk selection strategy in the encoder.
In this work, we choose to apply the topk selection on both the encoder and decoder for the consideration of the speed.
By adopting the topk selection strategy, the generated mapping matrices are guaranteed to be sparse, which can be leveraged to accelerate both the training and the inference of the model.

\setlength{\tabcolsep}{4pt}
\begin{table}
\begin{center}
\caption{Reconstruction errors with different mapping matrices }
\label{table_prior_map}
\begin{tabular}{l|ccc}
method & QEM & No fusing & Fusing              \\\hline
Error   & 0.939 &0.693 & 0.519
\end{tabular}
\end{center}
\end{table}

\subsection{Effect of fusing mapping matrices}
We also study the effect of fusing learned mapping matrices with precomputed mapping matrices. The results are shown in Table~\ref{table_prior_map}. By using the learned mapping matrices only, we can also significantly reduce the reconstruction error. Combining both mapping matrices by a linear fusion, we can further lower the reconstruction error. This is possible due to that the precomputed mapping matrices can benefit the training of the other components, including the convolution and the fully connected layers, especially at the early stage of the training phase.

\section{Parameter sensitivity studies}\label{sec_sup_exp_parameter_sensitivity}
In this section, we provide parameter sensitivity studies to get better understanding of the proposed feature aggregation module.
The experiments are conducted on COMA dataset by using the spectral convolution with the latent dimension as 8.

\subsection{Initialization of the weight $w_a$}
\begin{figure}[tb]
\centering
\includegraphics[width=.8\linewidth]{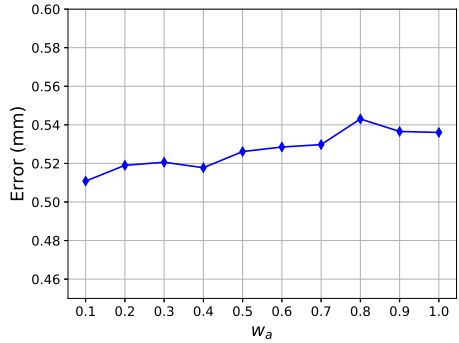}
\caption{Reconstruction errors with different initialization value for the weight $w_a$}
\label{fig_coma_weight}
\end{figure}

In Fig.~\ref{fig_coma_weight}, we show the reconstruction errors by training the model with different initialization values for the weight $w_a$.
While a smaller weight can lead to better performance, the performance variation is not notable.

\begin{figure}[tb]
\centering
\includegraphics[width=.8\linewidth]{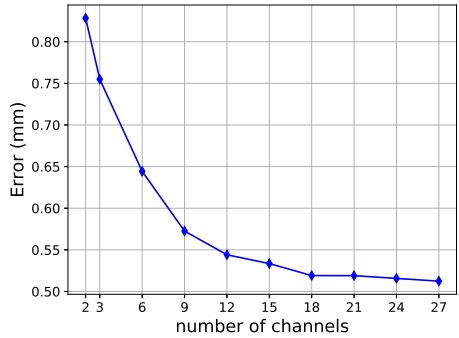}
\caption{Reconstruction errors with different number of channels for the keys and queries in the feature aggregation module}
\label{fig_coma_channel}
\end{figure}

\subsection{Number of channel $c$}
In Fig.~\ref{fig_coma_channel}, we show the variation of the reconstruction error with respect to the dimension of channels of the keys and queries in the feature aggregation module.
By increasing $c$ from 2 to 18, we can observe a significant drop in the corresponding error.
The performance is almost saturated when $c$ is larger than 18.
Note that we can surpass the QEM method even setting $c=2$ for our feature aggregation module.

\subsection{Top $k$ for the encoder and decoder}

\begin{figure}[tb]
\centering
\includegraphics[width=.8\linewidth]{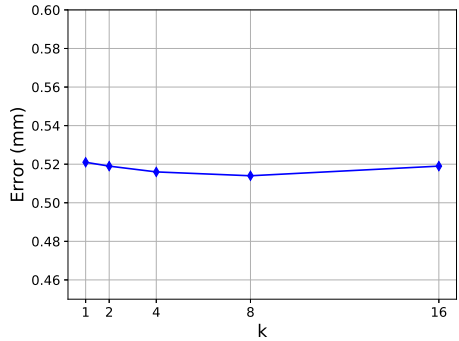}
\caption{Reconstruction errors with different $k$ for topk selection in the encoder}
\label{fig_coma_topk_enc}
\end{figure}

\begin{figure}[tb]
\centering
\includegraphics[width=.8\linewidth]{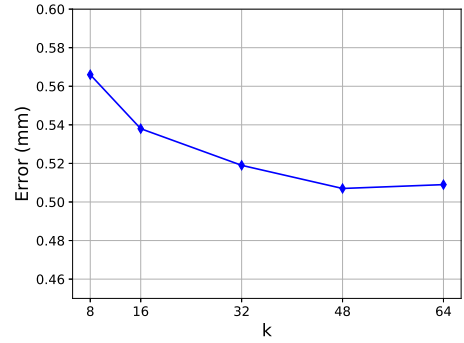}
\caption{Reconstruction errors with different $k$ for topk selection in the decoder }
\label{fig_coma_topk_dec}
\end{figure}

In Figs.~\ref{fig_coma_topk_enc} and~\ref{fig_coma_topk_dec}, we show the variation of the reconstruction error with respect to the $k$ value in the encoder and decoder, respectively.
By changing the $k$ value in the encoder, the performance is only slightly influenced.
In contrast, larger $k$ in the decoder would lead to better generalization with lower error.

\subsection{Initialization of the keys and queries}
We also study the effect of different initialization schemes of the keys and queries.
Table~\ref{table_init} gives the reconstruction errors of three different initialization, namely normal, uniform and template.
In the normal and uniform settings, we initialize the keys and queries by the random normal and uniform distributions, respectively.
In the precomputed setting, we use the vertex positions at each level computed by the mesh decimation to initialize the keys and queries. This is the initialization we adopt in this paper.
The precomputed based initialization outperforms the others significantly.
\setlength{\tabcolsep}{4pt}
\begin{table}
\begin{center}
\caption{Reconstruction errors with different initializations for the keys and queries }
\label{table_init}

\begin{tabular}{l|ccc}
method      & normal    & uniform       & precomputed    \\\hline
error (mm)  & 0.649     & 0.619         & 0.519
\end{tabular}
\end{center}
\end{table}

\section{Complexity}\label{sec_sup_discussion}
By directly parameterizing the mapping matrices, the complexity is with quadric scale $\mathcal{O}(n_{l}n_{l-1})$.
By using our attention based mechanism, the complexity to model the mapping matrices is reduced to a linear scale $\mathcal{O}(n_{l}+n_{l-1})$ which  parameterizes the keys and queries.
Thus, we provide a feasible solution to circumvent the over-parameterziation problem.

\end{document}